\documentclass[a4paper]{article}

\usepackage[utf8x]{inputenc}
\usepackage[T1]{fontenc}

\usepackage[a4paper,margin=1in]{geometry}

\usepackage[numbers]{natbib}
\usepackage[usenames,dvipsnames]{xcolor}

\definecolor{DarkRed}{rgb}{0.368,0.097,0.078}
\definecolor{DarkBlue}{rgb}{0.2,0.2,0.6}

\usepackage[colorlinks = true,
    linkcolor = blue,
    anchorcolor = blue,
    citecolor = blue,
    filecolor = blue,
    urlcolor = DarkRed,
    pagebackref]{hyperref}

\usepackage{amsthm} 
\usepackage{mathtools,thmtools}

\usepackage{enumitem}
\usepackage{bm}
\usepackage{xspace}
\usepackage{nicefrac}

\usepackage[capitalise]{cleveref}
\usepackage{dsfont}

\declaretheoremstyle[
	    spaceabove=\topsep, 
	    spacebelow=\topsep, 
	    headfont=\normalfont\bfseries,
	    bodyfont=\normalfont\itshape,
	    notefont=\normalfont\bfseries,
	    notebraces={(}{)},
	    postheadspace=0.5em, 
	    headpunct={},
	    postfoothook=\noindent\ignorespaces
    ]{theorem}
\declaretheorem[style=theorem,numberwithin=section]{theorem}

\declaretheoremstyle[
	    spaceabove=\topsep, 
	    spacebelow=\topsep, 
	    headfont=\normalfont\bfseries,
	    bodyfont=\normalfont,
	    notefont=\normalfont\bfseries,
	    notebraces={(}{)},
	    postheadspace=0.5em, 
	    headpunct={},
	    postfoothook=\noindent\ignorespaces
    ]{definition}

\declaretheoremstyle[
        spaceabove=\topsep, 
        spacebelow=\topsep, 
        headfont=\normalfont\bfseries,
        bodyfont=\normalfont,
        notefont=\normalfont\bfseries,
        notebraces={}{},
        postheadspace=0.5em, 
        qed=$\blacksquare$, 
        headpunct={},
        postfoothook=\noindent\ignorespaces
    ]{proofstyle}
\declaretheorem[style=proofstyle,numbered=no,name=Proof]{proof}

\declaretheorem[style=theorem,sibling=theorem,name=Lemma]{lemma}

\declaretheorem[style=theorem,numbered=no,name=Theorem]{theorem*}
\declaretheorem[style=theorem,numbered=no,name=Lemma]{lemma*}
\declaretheorem[style=theorem,numbered=no,name=Corollary]{corollary*}
\declaretheorem[style=theorem,numbered=no,name=Proposition]{proposition*}
\declaretheorem[style=theorem,numbered=no,name=Claim]{claim*}
\declaretheorem[style=theorem,numbered=no,name=Fact]{fact*}
\declaretheorem[style=theorem,numbered=no,name=Observation]{observation*}
\declaretheorem[style=theorem,numbered=no,name=Conjecture]{conjecture*}

\declaretheorem[style=definition,sibling=theorem,name=Definition]{definition}
\declaretheorem[style=definition,sibling=theorem,name=Remark]{remark}

\declaretheorem[style=definition,numbered=no,name=Definition]{definition*}
\declaretheorem[style=definition,numbered=no,name=Remark]{remark*}
\declaretheorem[style=definition,numbered=no,name=Example]{example*}
\declaretheorem[style=definition,numbered=no,name=Question]{question*}

\DeclareMathAlphabet{\mathbfsf}{\encodingdefault}{\sfdefault}{bx}{n}

\DeclareMathOperator*{\argmin}{arg\!\min}

\newcommand{\lr}[1]{\mathopen{}\left(#1\right)}
\newcommand{\Lr}[1]{\mathopen{}\big(#1\big)}

\newcommand{\lrbra}[1]{\mathopen{}\left[#1\right]}
\newcommand{\Lrbra}[1]{\mathopen{}\big[#1\big]}
\newcommand{\LRbra}[1]{\mathopen{}\Big[#1\Big]}

\newcommand{\lrnorm}[1]{\mathopen{}\left\|#1\right\|}

\newcommand{\lrset}[1]{\mathopen{}\left\{#1\right\}}

\newcommand{\LRset}[1]{\mathopen{}\Big\{#1\Big\}}

\newcommand{\lrabs}[1]{\mathopen{}\left|#1\right|}
\newcommand{\Lrabs}[1]{\mathopen{}\big|#1\big|}
\newcommand{\LRabs}[1]{\mathopen{}\Big|#1\Big|}

\usepackage{amsfonts,amsmath,amssymb}
\usepackage{mathtools}
\usepackage{float}
\usepackage{enumitem}
\usepackage{algorithm}
\usepackage{algpseudocode}

\newcommand{\cD}{\mathcal{D}}

\newcommand{\cO}{\mathcal{O}}
\newcommand{\cK}{\mathcal{K}}

\newcommand{\cM}{\mathcal{M}}
\newcommand{\cH}{\mathcal{H}}
\newcommand{\cA}{\mathcal{A}}
\newcommand{\cX}{\mathcal{X}}
\newcommand{\cY}{\mathcal{Y}}
\newcommand{\cZ}{\mathcal{Z}}
\newcommand{\cF}{\mathcal{F}}

\newcommand{\cG}{\mathcal{G}}
\newcommand{\cU}{\mathcal{U}}

\newcommand{\cN}{\mathcal{N}}
\newcommand{\cP}{\mathcal{P}}

\newcommand{\hR}{\mathbb{R}}
\newcommand{\hE}{\mathbb{E}}

\newcommand{\hP}{\mathbb{P}}
\newcommand{\hI}{\mathbb{I}}

\newcommand{\err}{\mathrm{Err}}
\newcommand{\fat}{\mathrm{fat}}
\newcommand{\RERM}{\mathrm{RERM}}
\newcommand{\ERM}{\mathrm{ERM}}
\newcommand{\medboost}{\mathrm{MedBoost}}
\newcommand{\med}{\mathrm{Med}}
\newcommand{\pac}{\mathrm{PAC}}

\newcommand{\RE}{\mathrm{RE}}
\newcommand{\AG}{\mathrm{AG}}
\newcommand{\VC}{\mathrm{VC}}

\title{Adversarially Robust PAC Learnability of Real-Valued Functions}

\author{
    Idan Attias \thanks{Department of Computer Science, Ben-Gurion University; \texttt{idanatti@post.bgu.ac.il}.} 
    \and  Steve Hanneke\thanks{Department of Computer Science, Purdue University; \texttt{steve.hanneke@gmail.com.}}
}

\begin{document}
\maketitle

\begin{abstract}
We study robustness to test-time adversarial attacks in the regression
setting with $\ell_p$ losses 
and arbitrary perturbation sets. We address the
question of which function classes are PAC learnable in
this setting.  We show that classes of finite fat-shattering dimension
are learnable in both realizable and agnostic settings.  Moreover, for convex function classes, they are even
properly learnable.  In contrast, some non-convex function classes
provably require improper learning algorithms.    
Our main technique is based on a
construction of an adversarially robust sample compression scheme of a
size determined by the fat-shattering dimension. Along the way, we introduce a novel agnostic sample compression scheme for real-valued functions, which may be of independent interest.
\end{abstract}
\section{Introduction}
Learning a predictor that is resilient to test-time adversarial attacks is a fundamental problem in contemporary machine learning. A long line of research has studied the vulnerability of deep learning-based models to small perturbations of their inputs (e.g., \citet{szegedy2013intriguing,biggio2013evasion,goodfellow2014explaining,madry2017towards}). From the theoretical standpoint, there has been a lot of effort to provide provable guarantees of such methods (e.g., \citet{feige2015learning,schmidt2018adversarially,khim2018adversarial,yin2019rademacher,cullina2018pac,attias2019improved,attias2022characterization,montasser2021transductive,montasser2020efficiently,montasser2020reducing,montasser2021adversarially,ashtiani2020black,dan2020sharp,awasthi2020adversarial,awasthi2021existence,awasthi2021calibration,awasthi2022h,awasthi2022multi,awasthi2023theoretically,bhattacharjee2021sample,xing2021adversarially,mao2023cross}), which is the focus of this work.

In the robust $\pac$ learning framework, the problem of learning binary function classes was studied by \citet{montasser2019vc}. They showed that uniform convergence does not hold in this setting, and as a result, robust empirical risk minimization is not sufficient to ensure learnability. Yet, they showed that $\VC$ classes are learnable, by considering an improper learning rule; the learning algorithm outputs a function that is not in the function class that we aim to learn.

In this work, we provide a theoretical understanding of the robustness of real-valued predictors in the $\pac$ learning model, with arbitrary perturbation sets.
The work of \citet{attias2019improved} considered this question for finite perturbation sets, they obtained sample complexity guarantees based on uniform convergence, which is no longer true for arbitrary perturbation sets.
We address the fundamental question, \textit{which real-valued function classes are robustly learnable?} 

Furthermore, we study the robust learnability of convex classes,
a natural and commonly studied subcategory for regression.
We address the question,
\textit{are real-valued convex classes properly robustly learnable?} 
On the one hand, some non-convex function classes provably require improper learning due to \citet{montasser2019vc}. On the other hand, \citet{mendelson2017optimal} showed that non-robust regression with the mean squared error is properly learnable.

We study the following learning models for real-valued functions. An adversarial attack is formalized by a perturbation function $\cU:\cX\rightarrow 2^\cX$, where $\cU(x)$ is the set of possible perturbations (attacks) on $x$. In practice, we usually consider $\cU(x)$ to be the $\ell_1$ ball centered at $x$. 
In this work, we have no restriction on $\cU$, besides $x\in\cU(x)$.
Let $\cD$ be an unknown distribution over $\cX\times [0,1]$ and let  $\cH\subseteq [0,1]^\cX$ be a concept class.
In our first model, the robust error of concept $h$ is defined as 
\begin{align*}
    \err_{\ell_p}(h;\cD)
    =
    \hE_{(x,y)\sim \cD}\lrbra{\sup_{z\in\cU(x)}\lrabs{h(z)- y}^p},
    1\leq p \leq \infty.
\end{align*}
The learner gets an i.i.d. sample from $\cD$, and would like to output function $\hat{h}$, such that with high probability,
\begin{align}\label{eq:small-error}
    \err_{\ell_p}(\hat{h};\cD)
    \leq
    \inf_{h\in \cH}\err_{\ell_p}(h;\cD) + \epsilon.
\end{align}
The sample complexity for learning $\cH$ is the size of a minimal i.i.d. sample from $\cD$ such that there exists a learning algorithm with output as in \cref{eq:small-error}.
We refer to this model as \emph{Robust Regression} with $\ell_p$ robust loss. This is a robust formulation of the classic nonparametric regression setting.

In the second model, the robust error of concept $h$ is defined as 
\begin{align*}
    \err_\eta(h;\cD)
    =
    \hE_{(x,y)\sim \cD}\lrbra{\hI \lrset{\sup_{z\in\cU(x)}\lrabs{h(z)- y} \geq \eta}}.
\end{align*}
We refer to the loss function in this model as \emph{cutoff loss}, where $\eta > 0$ is a predefined cutoff parameter.
 The learner gets an i.i.d. sample from $\cD$, and would like to output function $\hat{h}$, such that with high probability,
 \begin{align*}
    \err_{\eta+\beta}(\hat{h};\cD)
    \leq
    \inf_{h\in\cH} \err_{\eta}(h;\cD)+\epsilon,
\end{align*}
 where $\beta > 0$ is a predefined parameter. The sample complexity is defined similarly to the previous model.
 We refer to this model as \emph{Robust} $(\eta,\beta)$\emph{-Regression}.
The non-robust formulation of this setting was studied by, e.g.,
\citet{anthony2000function,simon1997bounds}. 
See also \citet[section 21.4]{anthony1999neural} and references therein.

\paragraph{Main results.} 
Denote the $\gamma$-fat-shattering dimension of $\cH$ by $\fat\lr{\cH,\gamma}$, and the dual $\gamma$-fat-shattering dimension by $\fat^*\lr{\cH,\gamma}$, which is the dimension of the dual class. The dimension of the dual class is finite as long as the $\gamma$-fat-shattering of the primal class is finite (see \citet{kleer2021primal}, and \cref{eq:dual-fat}).
\begin{itemize}[leftmargin=0.4cm]
    \item In \cref{sec:lp-regression}
    we provide a learning algorithm for robust regression with $\ell_p$ losses,
    with sample complexity \footnote{$\Tilde{\cO}$ hides polylogarithmic factors in the specified expression.}
        \begin{align*}
        \Tilde{\cO}\lr{\frac{\fat^3\lr{\cH,\epsilon/p}{\fat^*\lr{\cH,\epsilon/p}}}{\epsilon^5}}
        .
    \end{align*}
    Moreover, this algorithm is \emph{proper} for convex function classes.
    We circumvent a negative result regarding non-convex function classes, for which proper learning is impossible, even for binary-valued functions \citep{montasser2019vc}.

    \item In \cref{sec:l1-regression} we provide a learning algorithm with a substantial sample complexity improvement,
        \begin{align*}
        \Tilde{\cO}\lr{\frac{\fat\lr{\cH,\epsilon/p}{\fat^*\lr{\cH,\epsilon/p}}}{\epsilon^2}}.
    \end{align*}
    \item In \cref{sec:eta-beta-regression}, we provide learning algorithms for the $(\eta,\beta)$-robust regression setting in the realizable and agnostic settings. Our sample complexity for the realizable case is 
    \begin{align*}
        \Tilde{\cO}\lr{\frac{\fat\lr{\cH,\beta}{\fat^*\lr{\cH,\beta}}}{\epsilon}},
    \end{align*}
     and
        \begin{align*}
        \Tilde{\cO}\lr{\frac{\fat\lr{\cH,\beta}{\fat^*\lr{\cH,\beta}}}{\epsilon^2}}
    \end{align*}
    for the agnostic case.
\end{itemize}
\paragraph{Technical contributions and related work.}
The setting of agnostic adversarially robust regression with finite perturbation sets was studied by
\citet{attias2019improved}. Subsequently, improved bounds appeared in \citet{kontorovich2021fat}.
Adversarially robust $\pac$ learnability of binary-valued function classes with arbitrary perturbation sets was studied by \citet{montasser2019vc}. They showed that uniform convergence does not hold in this setting, which means that some classes provably require improper learning. Their main technique is constructing a sample compression scheme from a boosting-style algorithm, where the generalization follows from sample compression bounds.

First, we explain our new technical ideas behind the algorithms for robust $(\eta,\beta)$-regression, and compare it to the ones of \citet{montasser2019vc} in the classification setting.
We then explain why the approach for learning these models fails in the general robust regression setting and introduce the new ingredients 
behind the proofs for this setting.

\textit{Robust $(\eta,\beta)$-regression.} 
We construct an adversarially robust sample compression scheme of a size determined by the fat-shattering dimension of the function class. The following steps are different from the binary-valued case.
First, we use a robust
\textit{boosting algorithm for real-valued functions}.
In the non-robust setting, \citet{hanneke2019sample} showed how to convert a boosting algorithm (originally introduced by \citet{kegl2003robust}), into a sample compression scheme.
In order to find weak learners (and prove their existence), we rely on
\textit{generalization from approximate interpolation} (see \citet{anthony2000function} and \citet[section 21.4]{anthony2000function}). The idea is that any function $f\in\cF$ that approximately interpolates a sample $S\sim \cD^m$, that is, $\lrabs{f(x)-y}\leq \eta$ for $(x,y)\in S$, also satisfies that $\cP\lrset{(x,y):\lrabs{f(x)-y} \leq \eta+\beta} > 1-\epsilon$ with high probability, as long as $\Tilde{\cO}\lr{\fat(\cF,\beta)/\epsilon}\leq |S|$. Crucially, this result relies on uniform convergence and does not apply to the robust loss function.
Another difference is 
 in the discretization step. In the classification setting, we inflate the data set to include all possible perturbations (potentially infinite set). We then define a function class $\hat{\cH}$ by running a robust empirical minimizer on every subset of size $\VC(\cH)$ from the training set, where $\cH$ is the class we want to learn. $\hat{\cH}$ induces a finite partition on the inflated set into regions, such that any $h\in\hat{\cH}$ has a constant error in each region.
 This is no longer true in the real-valued case. Instead, we discretize the inflated set by taking a \emph{uniform cover} using the supremum metric and controlling the errors that arise from the cover.

\textit{Robust regression.} We first explain which natural techniques fail.
We cannot run boosting for the $\ell_p$ loss as explained by \citet{hanneke2019sample}:
"\citet[Remark 2.1]{duffy2002boosting}
 spell out a central technical challenge: no boosting algorithm can always force the base regressor to output a useful function by simply modifying the distribution over the sample. This is because unlike a binary classifier, which localizes errors on specific examples, a real-valued hypothesis can spread its error evenly over the entire sample and it will not be affected by reweighting". 

As a first attempt, we could try to learn with respect to the cutoff loss (with a fixed cutoff parameter), and conclude learnability in the general regression setting.
 However, the $\ell_p$ loss can spread over different values for different points, which means that this approach fails.
 In another possible attempt, we could try to solve the realizable case first and try to reduce agnostic to realizable learning as in \citet{montasser2019vc} for binary-valued functions, as we prove the agnostic setting for robust $(\eta,\beta)$-regression. However, this attempt fails for the same reasons we mentioned above.
 
 Therefore, we introduce a novel technique for handling \emph{changing} cutoffs. We establish generalization from approximate interpolation with \emph{different} cutoff parameters, and thereby, we find a learner that approximates the loss of the target function on different points.
 Utilizing this idea, we provide a learning algorithm for $\ell_p$ robust loss that constructs an ensemble and predicts with the average. Further, we show that this algorithm is \emph{proper} for convex function classes.  In contrast, some non-convex function classes provably require improper learning \citep{montasser2019vc}.
Moreover, we show how to reduce the sample complexity substantially with a different algorithm, by constructing an ensemble of weak learners and predicting with the median. 
 Both algorithms can be represented as an agnostic sample compression scheme for the robust loss. This is a new result since constructing a sample compression scheme for real-valued functions is known only for the realizable setting \cite{hanneke2019sample}. We believe that this technique may be of independent interest.
\section{Problem Setup and Preliminaries}\label{sec:prelim}
Let $\cH\subseteq[0,1]^{\cX}$ be a concept class. We implicitly assume that all concept classes are satisfying mild measure-theoretic conditions (see e.g., \citet[section 10.3.1]{dudley1984course} and \citet[appendix C]{pollard2012convergence}).
Let $\cD$ be a distribution over $\cX\times \cY$, where $\cY=[0,1]$.
Define a perturbation function $\cU: \cX \rightarrow 2^{\cX}$ that maps an input to an arbitrary set $\cU(x)\subseteq \cX$, such that $x\in \cU(x)$. 

We consider the following loss functions. 
For $1\leq p \leq\infty$, define the $\ell_p$ \textit{robust} \textit{loss} function of $h$ on $(x,y)$ with respect to a perturbation function $\cU$,
\begin{align}\label{def:lp-loss}
    \ell_{p,\cU}(h;(x,y))=\sup_{z\in\cU(x)}\lrabs{h(z)- y}^p.
\end{align}
We define also the
$\eta$\textit{-ball} \textit{robust loss} function of $h$ on $(x,y)$ with respect to a perturbation function $\cU$, 
\begin{align}\label{def:eta-loss}
  \ell^\eta_{\cU}(h;(x,y))=\hI \lrset{\sup_{z\in\cU(x)}\lrabs{h(z)- y} \geq \eta}.
\end{align}
The non-robust version of this loss function is also known as $\eta$-ball or $\eta$-tube loss (see for example \citet[Section 21.4]{anthony1999neural}).

Define the error of a function $h$ on distribution $\cD$, with respect to the $\ell_p$ robust loss,
\begin{align*}
    \err_{\ell_p}(h;\cD) 
    =
    \hE_{(x,y)\sim \cD}\lrbra{\sup_{z\in\cU(x)}\lrabs{h(z)- y}^p},
\end{align*}
and the error with respect to the  $\eta$-ball robust loss
\begin{align*}
    \err_\eta(h;\cD)
    =
    \hE_{(x,y)\sim \cD}\lrbra{\hI \lrset{\sup_{z\in\cU(x)}\lrabs{h(z)- y} \geq \eta}}.
\end{align*}

Note that in our model the learner is tested on the original label $y$ while observing only the perturbed example $z$. There are formulations of robustness where the learner is compared to the value of the optimal function in the class on the perturbed example, i.e., if the optimal function in the class is $h^\star$, then the $\ell_1$ robust loss would be $\sup_{z\in\cU(x)}\lrabs{h(z)- h^\star(z)}$. 
For more details and comparisons of the two models, see \citet{gourdeau2021hardness,diochnos2018adversarial,bubeck2019adversarial}.
\paragraph{Learning models.} We precisely define the models for robustly learning real-valued functions. 
Our first model is learning with the $\ell_p$ robust loss (see \cref{def:lp-loss}), we refer to this model as \emph{Robust Regression}.

\begin{definition}[Robust regression]\label{def:rob-reg}
For any $\epsilon,\delta\in (0,1)$, the sample complexity robust $(\epsilon,\delta)$-$\pac$ learning a concept class $\cH\subseteq [0,1]^\cX$ with the $\ell_p$ robust loss,
denoted by $\cM(\epsilon,\delta,\cH,\cU,\ell_p)$, is the smallest integer $m$ such that the following holds: there exists a learning algorithm $\cA :\lr{\cX\times\cY}^m\rightarrow [0,1]^{\cX}$, such that for any distribution $\cD$ over $\cX\times [0,1]$,
for an i.i.d. random sample $S \sim \cD^m$, with probability at least $1-\delta$ over $S$, it holds that
\begin{align*}
      \err_{\ell_p}\lr{\cA(S);\cD}   
    \leq
      \inf_{h\in \cH}\err_{\ell_p}\lr{h;\cD}+\epsilon.
\end{align*}
    If no such $m$ exists, define $\cM(\epsilon,\delta,\cH,\cU,\ell_p) = \infty$,
    and $\cH$ is not robustly $(\epsilon,\delta)$-$\pac$ learnable.
We use the shorthand
$\cM=\cM(\epsilon,\delta,\cH,\cU,\ell_p)$
for notational simplicity.
\end{definition}

Our second model is learning with the $\eta$-ball robust loss (see \cref{def:eta-loss}) in the realizable and agnostic settings, we refer to this model by \textit{Robust $(\eta,\beta)$-regression}.
We say that a distribution $\cD$ is $\eta$\textit{-uniformly realizable} with respect to $\cH$ and $\cU$, if there exists $h^\star\in\cH$
such that
\begin{align}\label{def:eta-uni-re}
\err_\eta(h^\star;\cD)=0.
\end{align}

\begin{definition}[Robust $(\eta,\beta)$-regression]\label{def:rob-uni-reg}
For any $\eta,\beta,\epsilon,\delta\in (0,1)$, the sample complexity of \textit{realizable} robust $(\eta,\beta,\epsilon,\delta)$-$\pac$ learning a concept class $\cH\subseteq [0,1]^\cX$, 
denoted by $\cM_{\RE}(\eta,\beta,\epsilon,\delta,\cH,\cU)$, is the smallest integer $m$ such that the following holds: there exists a learning algorithm $\cA :\lr{\cX\times\cY}^m\rightarrow [0,1]^{\cX}$, such that for any distribution $\cD$ over $\cX\times [0,1]$ that is $\eta$-uniformly realizable w.r.t. $\cH$ and $\cU$ (see \cref{def:eta-uni-re}), for an i.i.d. random sample $S \sim \cD^m$, with probability at least $1-\delta$ over $S$, it holds that 
\begin{align*}
      \err_{\eta+\beta}\lr{\cA(S);\cD}
      \leq
      \epsilon.
\end{align*}
    If no such $m$ exists, define $\cM_{\RE}(\eta, \beta, \epsilon,\delta,\cH,\cU) = \infty$,
    and $\cH$ is not robustly $(\eta,\beta,\epsilon,\delta)$-$\pac$ learnable.

The \textit{agnostic} sample complexity, denoted by $\cM_{\AG}(\eta,\beta,\epsilon,\delta,\cH,\cU)$, is defined similarly with the following difference. We require the learning algorithm to output a function, such that with probability at least $1-\delta$,
\begin{align*}
      \err_{\eta+\beta}\lr{\cA(S);\cD} 
      \leq \inf_{h\in \cH}\err_{\eta}\lr{h;\cD} + \epsilon.
\end{align*}
\end{definition}
We use the shorthand
$\cM^{\eta,\beta}_{\RE}=\cM_{\RE}(\eta,\beta,\epsilon,\delta,\cH,\cU)$ and
$\cM^{\eta,\beta}_{\AG}=\cM_{\AG}(\eta,\beta,\epsilon,\delta,\cH,\cU)$ for notational simplicity.

We do not define the setting of robust regression in the realizable setting since it coincides with the realizable setting of robust $(\eta,\beta)$-regression, by taking $\eta=0, \beta=\epsilon/2$, and re-scaling $\epsilon$ to $\epsilon/2$. 
Moreover, we could define the $\ell_p$ variant of the $\eta$-ball loss in robust $(\eta,\beta)$-regression, however, results for our definition translate immediately by taking $\eta^{1/p}$.

Note that there is a fundamental difference between the models.
In the robust $(\eta,\beta)$-regression, we demand from the learning algorithm to find a function that is almost everywhere within $\eta+\beta$ from the target function in class.
That is, on $1-\epsilon$ mass of elements in the support of $\cD$, we find an approximation up to $\eta+\beta$. On the other hand, in the robust regression model, we aim to be close to the target function on average, and the error can possibly spread across all elements in the support.

\textit{Proper and improper learning algorithms}. The learning algorithm is not limited to returning a function that is inside the concept class that we aim to learn. When learning a class $\cH$, whenever the learning algorithm returns a function inside the class, that is, $\cA :\lr{\cX\times\cY}^m\rightarrow \cH$, we say that the algorithm is proper and the class in properly learnable. Otherwise, we say that the algorithm is improper. Improper
algorithms are extremely powerful and using them often circumvents computational issues and sample complexity barriers \citep{srebro2004maximum,candes2012exact,anava2013online,hazan2015classification,hanneke2016optimal,hazan2016non,hazan2012near,agarwal2019online,daniely2011multiclass,daniely2014optimal,angluin1988queries,montasser2019vc}.
\paragraph{Oracles.} 
We rely on the following robust empirical risk minimizers.
Let a set $S=\lrset{\lr{x_i,y_i}}^m_{i=1}$. Define an $\epsilon$-approximate $\psi$-robust empirical risk minimizer
$\psi\text{-}\RERM_\cH: \lr{\cX\times\cY}^m\times [0,1]^m\times(0,1) \rightarrow \cH$, 
\begin{align}\label{def:eta-rerm}
\begin{split}
    \psi\text{-}\RERM_\cH(S,\psi|_S,\epsilon) 
    :=
    \argmin_{h\in\cH}\frac{1}{m}\sum_{(x,y)\in S}\hI\lrbra{ \sup_{z\in\cU(x)}\lrabs{h(z)-y} \geq \psi(x,y)+\epsilon},
\end{split}
\end{align}
where $\psi|_S=\lr{\psi(x_1,y_1),\ldots,\psi(x_m,y_m)}$. We refer to $\psi(x,y)$ as \emph{cutoff} parameters.
Note that $\psi$ is a function of $(x,y)$ and not necessarily a constant.

For $p\in [1,\infty)$, define the empirical robust loss of a function $h$ on a labeled sample $S$ by  
\begin{align*}
    L_p(h,S,\cU)
    =
    \frac{1}{m}\sum_{(x,y)\in S}
\sup_{z\in\cU(x)}\lrabs{h(z)-y}^p,
\end{align*}
and for $p=\infty$
\begin{align*}
    L_\infty(h,S,\cU)
    =
    \max_{(x,y)\in S}
    \sup_{z\in\cU(x)}\lrabs{h(z)-y}.
\end{align*}
For the empirical non-robust loss, we omit $\mathcal{U}$ from the notation and use $L_p(h,S)$.

Define a robust empirical risk minimizer for the $\ell_p$ robust loss, $\ell_p\text{-}\RERM_\cH: \lr{\cX\times\cY}^m\rightarrow \cH$ by
\begin{align}\label{def:lp-rerm}
\ell_p\text{-}\RERM_\cH(S) 
:= 
\argmin_{h\in\cH}L_p(h,S,\cU).
\end{align}
%
\paragraph{Complexity measures.}
\textit{Fat-shattering dimension.}
Let $\cF\subseteq[0,1]^\cX$ and $\gamma > 0$. We say that $S = \{x_1, \ldots, x_m\} \subseteq \cX$ is $\gamma$-shattered by $\cF$ if there exists a witness $r = (r_1, \ldots,r_m) \in [0,1]^m$ such that for each $\sigma = (\sigma_1, \ldots, \sigma_m) \in \{-1, 1\}^m$ there is a function $f_{\sigma} \in \cF$ such that
\[
\forall i \in [m] \; 
\begin{cases}
	f_{\sigma}(x_i) \geq r_i + \gamma, & \text{if $\sigma_i = 1$}\\
    f_{\sigma}(x_i) \leq r_i - \gamma, & \text{if $\sigma_i = -1$.}
 \end{cases}
\]
The fat-shattering dimension of $\cF$ at scale $\gamma$, denoted by $\fat(\cF,\gamma)$, is 
the cardinality of the largest set of points in $\cX$ that can be $\gamma$-shattered by $\cF$.
This parametrized variant of the Pseudo-dimension \citep{alon1997scale}
was first proposed by \citet{kearns1994efficient}.
Its key role in learning theory
lies in characterizing the PAC learnability
of real-valued function classes
\citep{alon1997scale,bartlett1998prediction}.

\textit{Dual fat-shattering dimension. }Define the dual class $\cF^* \subseteq [0,1]^\cF$ of $\cF$ as the set of all functions $g_w: \cF \rightarrow [0,1]$ defined by $g_w(f) = f(w)$. If we think of a function class as a matrix whose rows and columns are indexed by functions and points, respectively, then the dual class is given by the transpose of the matrix. 
The dual fat-shattering at scale $\gamma$, is defined as the fat-shattering at scale $\gamma$ of the dual-class and denoted by $\fat^*\lr{\cF,\gamma}$.
We have the following bound due to \citet[Corollary 3.8 and inequality 3.1]{kleer2021primal},
\begin{align}\label{eq:dual-fat}
   \fat^*\lr{\cF,\gamma} 
   \lesssim 
   \frac{1}{\gamma}2^{\fat\lr{\cF,\gamma/2}+1}.
\end{align}
\textit{Covering numbers.}
We say that $\cG\subseteq[0,1]^\Omega$ is $\epsilon$-cover for $\cF\subseteq[0,1]^\Omega$ in $\lrnorm{\cdot}_\infty$ norm, if for any $f\in\cF$ the exists $g\in \cG$ such that for any $x\in \Omega$, $\lrabs{f(x)-g(x)}\leq \epsilon$. The $\epsilon$-covering number of $\cF$ is the minimal cardinality of any $\epsilon$-cover, and denoted by $\cN\lr{\epsilon,\cF,\lrnorm{\cdot}_\infty}$.

\paragraph{Approximate sample compression schemes.}
Following
\citet{david2016supervised},
a {\em selection scheme} $(\kappa,\rho)$ for a hypothesis
class $\cH\subset\cY^\cX$
is defined as follows.
A $k$-{\em selection} function $\kappa$
maps sequences \\
$\lrset{(x_1,y_1),\ldots,(x_m,y_m)}\in\bigcup_{\ell\ge1}\lrset{\cX\times\cY}^\ell$
to elements in
$\cK=\bigcup_{\ell\le k'}\lrset{\cX\times\cY}^\ell
\times
\bigcup_{\ell\le k''}\lrset{0,1}^\ell$,
where $k'+k''\le k$.
A {\em reconstruction} is a function $\rho:\cK\to\cY^\cX$.
We say that $(\kappa,\rho)$ is a $k$-size agnostic \textit{$\alpha$-approximate sample compression scheme} for $\cH$
if
$\kappa$ is a $k$-selection and
for all
$S = \lrset{(x_i,y_i):i\in [m]}$, and 
$f_S := \rho(\kappa(S))$
achieves $\cH$-competitive empirical loss:
\begin{align}\label{eq:exact-comression}
    L_p(f_S,S) &\le \inf_{h\in\cH}L_p(h,S)+\alpha.
\end{align}
Similarly, we define an agnostic \textit{$\alpha$-approximate adversarially robust sample compression scheme} if $f_S := \rho(\kappa(S))$
achieves $\cH$-competitive empirical robust loss:
\begin{align}\label{eq:exact-comression}
    L_p(f_S,S,\cU) &\le \inf_{h\in\cH}L_p(h,S,\cU)+\alpha.
\end{align}
When $p=\infty$ we refer to the sample compression as \textit{uniformly $\alpha$-approximate}.

\paragraph{Notation.} We use the notation $\Tilde{\cO}(\cdot)$ for omitting poly-logarithmic factors of \\
$\lr{\fat(\cH,\gamma),\fat^*(\cH,\gamma),1/\epsilon,1/\delta,1/\eta,1/\beta}$. We denote $[n]=\lrset{1,\ldots,n}$, and $\exp(\cdot)=e^{\lr{\cdot}}$.
$\lesssim$ and $\gtrsim$ denote inequalities up to a constant factor, and
$\approx$ denotes equality up to a constant factor. 
Vectors are written using bold symbols.

\section{Robust Regression}\label{sec:lp-regression}
In this section, we provide an algorithm and prove its sample complexity for robust regression with the $\ell_p$ loss. Moreover, our learning algorithm is \emph{proper} for convex function classes, arguably the most commonly studied subcategory of real-valued function classes for regression. This result circumvents a negative result from \citet{montasser2019vc}; there exist, non-convex function classes, where proper learning is impossible.

\begin{theorem}\label{thm:lp-regression} 
 \cref{alg:lp-regression-highvote} implies that the 
    sample complexity for robust $(\epsilon,\delta)$-PAC learning a concept class $\cH$ with the $\ell_p$ robust loss is 
    \[
\begin{dcases}
	\Tilde{\cO}\lr{\frac{\fat^3\lr{\cH,c\epsilon/p}{\fat^*\lr{\cH,c\epsilon/p}}}{\epsilon^5}+\frac{1}{\epsilon^2}\log \frac{1}{\delta}}, p\in [1,\infty)\\
 \Tilde{\cO}\lr{\frac{\fat^3\lr{\cH,c\epsilon}{\fat^*\lr{\cH,c\epsilon}}}{\epsilon^5}+\frac{1}{\epsilon^2}\log \frac{1}{\delta}}
   , p=\infty,
 \end{dcases}
\]
    for some numerical constant $c\in (0,\infty)$. 
Recall that $\fat^*\lr{\cF,\epsilon} 
   \lesssim 
   \frac{1}{\epsilon}2^{\fat\lr{\cF,\epsilon/2}+1}$ by \cref{eq:dual-fat}.
\end{theorem}

\begin{remark}
  The output of \cref{alg:lp-regression-highvote} is a convex combination of the functions from the concept class, which is a proper predictor, assuming convexity of the function class.  
\end{remark}

\begin{remark}
    Similar to non-robust regression, our results generalize to loss functions with bounded codomain $[0,M]$. The generalization bound should be multiplied by $pM^p$ and the scaling of the fat-shattering dimension should be $\epsilon/pM^p$.
\end{remark}

In the following result, we establish generalization from approximate interpolation for \emph{changing} cutoff parameters for different points.
This generalizes a result by \cite{anthony2000function}, where the cutoff parameter is fixed for all points. 
The proof is in \cref{app:generalization-interpolation}.
%

\begin{theorem}[Generalization from approximate interpolation with changing cutoffs]
\label{thm:generalization-interpolation-cutoffs}
Let $\cF\subseteq [0,1]^\cX$ be a function class with a finite fat-shattering dimension (at any scale). For any $\beta,\epsilon,\delta\in(0,1)$, any function $\psi:\cX\times \cY \rightarrow [0,1]$, any distribution $\cD$ over $\cX\times\cY$, for a random sample $S\sim \cD^m$, if 
$$
m=\cO\lr{\frac{1}{\epsilon}\lr{\fat\lr{\cF,\beta/8}\log^2\lr{\frac{\fat\lr{\cF,\beta/8}}{\beta\epsilon}}+\log\frac{1}{\delta}}},
$$ 
then with probability at least $1-\delta$ over $S$, for any ${f}\in\cF$ satisfying $\lrabs{{f}(x)-y}\leq \psi(x,y)+\beta$, $\forall{(x,y)\in S}$,
it holds that $\hP_{(x,y)\sim \cD}\lrset{(x,y): \lrabs{f(x)-y} \leq \psi(x,y)+2\beta}\geq 1- \epsilon$.
\end{theorem}

\begin{algorithm}[H]
    \caption{Improper Robust Regressor with High-Vote}\label{alg:lp-regression-highvote}
    \textbf{Input}: $\cH\subseteq [0,1]^\cX$, $S=\lrset{\lr{x_i,y_i}}^m_{i=1}$.\\
    \textbf{Parameters}: Approximation parameter $\epsilon$,  base learner sample size $d\geq 1$, number of Multiplicative Weights rounds $T\geq 1$, loss parameter $p\in [1,\infty]$.\\
    \textbf{Algorithms used:} $\ell_p\text{-}\RERM_\cH$ (\cref{def:lp-rerm}), $\psi\text{-}\RERM_\cH$ (\cref{def:eta-rerm}), a robust variant of $\mathrm{Multiplicative\; Weights}$ (\cref{alg:MW}).
    \begin{enumerate}
        \item Compute $h^\star\leftarrow \ell_p\text{-}\RERM_\cH(S)$. \\
            Denote $\psi(x,y)=\sup_{z\in\cU(x)}|h^\star(z)-y|$, $\forall (x,y)\in S$.
    	\item Inflate $S$ to $S_\cU$ to include all perturbed points. 
    	\item Discretize $\bar{S}_\cU\subseteq S_\cU$:
            (i) Construct a function class $\hat{\cH}$, where each $\hat{h}\in \hat{\cH}$ is obtained by $\psi\text{-}\RERM$ optimizer operating on $d$ points from $S$. The input cutoff parameters to the optimizer are $\psi(x,y)$, as computed in step 1.\\
            (ii) Let $\Tilde{\cH}=\hat{\cH}\cup \lrset{h^\star}$. Each $\lr{z,y}\in S_\cU$ defines a function in the dual space, $f_{\lr{z,y}}: \Tilde{\cH} \rightarrow \lrbra{0,1}$ such that $f_{\lr{z,y}}(h) = \Lrabs{h(z)- y}^p$. 
             Define $\bar{S}_\cU$ to be the minimal cover of $S_\cU$ at scale $\cO\lr{\epsilon/p}$ under the supremum norm.
    	\item Compute a robust $\mathrm{Multiplicative\; Weights}$ algorithm on $\bar{S}_\cU$. Let $\lrset{\hat{h}_1,\ldots,\hat{h}_T}$ be the returned set of classifiers.
    \end{enumerate}
    \textbf{Output:} $\hat{h}=\frac{1}{T}\sum^T_{i=1} \hat{h}_i$. 
\end{algorithm}

We construct an adversarially robust sample compression scheme of a size determined by the fat-shattering dimension of the function class. 
Recall that uniform convergence does not necessarily hold. Instead, we derive generalization from sample compression bounds. 
%
\paragraph{Proof overview and algorithm outline.} 
The complete proof is in \cref{app:lp-regression}.
We follow the steps in \cref{alg:lp-regression-highvote}.
\begin{enumerate} [leftmargin=0.4cm]
\item We start with computing a robust empirical risk minimizer (ERM) $h^\star$ on $S$ for the $\ell_p$ robust loss, $p\in [1,\infty]$. This defines the target loss we are aiming for at any point in $S$. In other words, the robust loss of $h^\star$ on $(x,y)$ defines a \emph{cutoff} $\psi(x,y)$ and our goal is to construct a predictor with a loss of $\psi(x,y)^p+\epsilon$ for any $(x,y)\in S$, which means that this predictor is an approximate robust ERM. In order to derive generalization, we cannot rely on uniform convergence. Instead, our predictor is based on a sample compression scheme from which we can generalize.

\item Inflate the training set by including all possible perturbations. Whenever the same perturbation is mapped to more than one input, we assign the label of the input with the smallest index to prevent ambiguity.
We denote this set by $S_\cU$. 

\item Discretize the set $S_\cU$ as follows: (i) Construct a set of functions $\hat{\cH}$, such that each function is the output of $\psi\text{-}\RERM$ for $\cH$ (defined in \cref{def:eta-rerm}), performing on a subset $S'\subseteq S$ of size 
\[
\begin{dcases}
    d \leftarrow \Tilde{\cO}\lr{\frac{1}{\epsilon}\fat\lr{\cH,c\epsilon/p}
    }, p\in [1,\infty)\\
   d\leftarrow
   \Tilde{\cO}\lr{\frac{1}{\epsilon}\fat\lr{\cH,c\epsilon}
   }, p=\infty.
\end{dcases}
\]
This means that for any $S'\subseteq S$ there exists $\hat{h}\in\hat{\cH}$ that is an approximate robust ERM on $S'$, that is, $\hat{h}$ is within $\psi(x,y)^p+\epsilon$ for any $(x,y)\in S'$.
The size of $\hat{\cH}$ is bounded $(m/d)^d$, where $\lrabs{S}=m$.

(ii) Let $\Tilde{\cH}=\hat{\cH}\cup \lrset{h^\star}$. Define a discretization $\bar{S}_\cU\subseteq S_\cU$ by taking a uniform cover of the dual space defined on $\Tilde{\cH}$.
In the dual space,
each $\lr{z,y}\in S_\cU$ defines a function $f_{\lr{z,y}}: \Tilde{\cH} \rightarrow \lrbra{0,1}$ such that $f_{\lr{z,y}}(h) = \Lrabs{h(z)- y}^p$. We take a minimal $\cO\lr{\epsilon/p}$-cover for $S_\cU$ with the supremum norm, which is of size $\cN\lr{\cO\lr{\epsilon/p},S_\cU,\lrnorm{\cdot}_\infty}$.
We use covering numbers arguments \citep{rudelson2006combinatorics} to upper bound the size of $\bar{S}_\cU$

\item Compute a variant of $\mathrm{Multiplicative\;Weights}$ (MW) update (\cref{alg:MW}) on $\bar{S}_\cU$ for $T \approx \log\lrabs{\bar{S}_\cU}$ rounds as follows. 
From \cref{thm:generalization-interpolation-cutoffs} and using the Lipschitzness of the $\ell_p$ loss, we know that for any distribution $\cP$ on $\bar{S}_\cU$, upon receiving an i.i.d. sample $S''$ from $\cP$ of size $d$, with probability $2/3$ over sampling $S''$ from $\cP$, for any ${h}\in\cH$ with 
$\forall (z,y)\in S'':\; \lrabs{{h}(z)-y}^p\leq \psi(z,y)^p+\epsilon$, 
it holds that \\
$\hP_{(z,y)\sim\cP}\lrset{(z,y): \lrabs{h(z)-y}^p\leq\psi(z,y)^p+2\epsilon}\geq 1-\epsilon$, where $\psi(z,y)$ is the $\psi(x,y)$ for which $z \in \cU(x)$.
We can conclude that for any distribution $\cP$ on $\bar{S}_\cU$, there exists such a set of points $S''\subseteq \bar{S}_\cU$. 
Then, we can find a set $S'$ of $d$ points in $S$ that 
$S''$ originated from. 
Formally, $S''\subseteq \bigcup_{(x,y)\in {S'}}\bigcup \lrset{(z,y): z\in \cU(x)}$.
We execute the optimizer $\hat{h}\leftarrow\psi\text{-}\RERM$ on $S'$ with the relevant cutoff parameters.
$\hat{h}$ has error of $\psi(z,y)^p+\epsilon$ on a fraction of $(1-\epsilon)$ points with respect to the distribution $\cP$.
We start with $\cP_1$ as the uniform distribution over $\bar{S}_{\cU}$ and find $\hat{h}_1$ respectively. We perform a multiplicative weights update on the distribution and find the next hypothesis w.r.t. the new distribution and so forth.
\end{enumerate}

Following the analysis of MW (or $\alpha$-Boost)
from \citet[Section 6]{schapire2013boosting}), we know that for any point in $\bar{S}_\cU$,
roughly $(1-\epsilon)$ base learners 
are within $\epsilon$ from the target cutoff. The rest $\epsilon$ fraction can contribute an error of at most $\epsilon$ since the loss is bounded by $1$. We get that for any point in $\bar{S}_\cU$, the average loss of hypotheses in the ensemble is within $3\epsilon$ from the target cutoff. Crucially, we use strong base learners in the ensemble.
By the covering argument, we get that for any point in $S_\cU$, the average loss of the ensemble is within $5\epsilon$,
\begin{align*}
\forall \lr{z,y}\in S_\cU:\;
\frac{1}{T}\sum^T_{i=1}\lrabs{\hat{h}_i(z)-y}^p
\leq
\psi\lr{z,y}^p+5\epsilon.
\end{align*}
We are interested that the average prediction $\frac{1}{T}\sum^T_{i=1} \hat{h}_i$ will be within the target cutoffs. For that reason,
we use the convexity of the $\ell_p$ loss to show that
\begin{align*}
    \lrabs{\frac{1}{T}\sum^T_{i=1}\hat{h}_i(z)-y}^p
    \leq
    \frac{1}{T}\sum^T_{i=1}\lrabs{\hat{h}_i(z)-y}^p.
\end{align*}
Therefore, we conclude that
\begin{align*}
    \forall \lr{z,y}\in S_\cU:\;
    \lrabs{\frac{1}{T}\sum^T_{i=1}\hat{h}_i(z)-y}^p
    \leq
    \psi(z,y)^p+5\epsilon,
\end{align*}
which implies that we have an approximate robust ERM for $S$,
\begin{align*}
    \forall \lr{x,y}\in S:\;
    \sup_{z\in\cU(x)}\lrabs{\frac{1}{T}\sum^T_{i=1}\hat{h}_i(z)-y}^p
    \leq
    \psi(x,y)^p+5\epsilon.
\end{align*}

The proof follows by applying a sample compression generalization bound in the agnostic case, bounding the compression size, and rescaling $\epsilon$.

For convex classes, we have a proper learner.
The output of the algorithm is a convex combination of functions from $\cH$ which is also in the class. 
\section{Improved Sample Complexity via Median  Boosting and Sparsification}\label{sec:l1-regression}
In this section, we provide an algorithm with a substantial sample complexity improvement.
The key technical idea in this result is to note that, if we replace base learners with weak learners in the improper ensemble predictor, we can still get an accurate prediction by taking the \emph{median} aggregation of the ensemble.
Thus, we incorporate a variant of median boosting for real-valued functions \citep{kegl2003robust,hanneke2019sample} in our algorithm.
Each base learner requires fewer samples and as a result, we improve the sample complexity.
On the contrary, in \cref{alg:lp-regression-highvote} we obtained accurate predictions for a $1-\cO\lr{\epsilon}$
 quantile of the predictors, and we output their average.

\begin{theorem}\label{thm:l1-regression}
     \cref{alg:l1-regression} implies that the sample complexity for robust $(\epsilon,\delta)$-PAC learning a concept class $\cH$ with the $\ell_p$ robust loss is 
       \[
\begin{dcases}
	\Tilde{\cO}\lr{\frac{\fat\lr{\cH,c\epsilon/p}{\fat^*\lr{\cH,c\epsilon/p}}}{\epsilon^2}+\frac{1}{\epsilon^2}\log \frac{1}{\delta}}, p\in [1,\infty)\\
 \Tilde{\cO}\lr{\frac{\fat\lr{\cH,c\epsilon}{\fat^*\lr{\cH,c\epsilon}}}{\epsilon^2}+\frac{1}{\epsilon^2}\log \frac{1}{\delta}}
   , p=\infty,
 \end{dcases}
\]
    for some numerical constant $c\in (0,\infty)$.
    Recall that $\fat^*\lr{\cF,\epsilon} 
   \lesssim 
   \frac{1}{\epsilon}2^{\fat\lr{\cF,\epsilon/2}+1}$ by \cref{eq:dual-fat}.
\end{theorem}

We shall define the notion of weak learners in the context of real-valued learners. 
\begin{definition}[Weak real-valued learner]\label{def:weak-learner}
Let $\xi\in(0,\frac{1}{2}]$, $\zeta\in [0,1]$.
We say that $f: \cX \rightarrow [0,1]$ is a $(\zeta,\xi)$-weak learner with respect to $\cD$ and a target function $h^\star\in\cH$ if
\begin{equation*}
\hP_{(x,y)\sim\cD}\lrset{(x,y):\lrabs{f(x)-y}>\lrabs{h^\star(x)-y}+\zeta}\leq
\frac{1}{2}-\xi.
\end{equation*}
\end{definition}
This notion of a weak learner must be formulated carefully. 
For example, taking a learner guaranteeing absolute loss at most $\frac{1}{2}-\xi$ is known to not be strong enough for boosting to work. On the other hand, by making the requirement too strong (for example, $\mathrm{AdaBoost.R}$ in \citet{freund1997decision}), then the sample complexity of weak learning will be high that weak learners cannot be expected to exist for certain function classes. 
We can now present an overview of the proof and the algorithm. 
\begin{algorithm}[H]
    \caption{Improper Robust Regressor}\label{alg:l1-regression}
    \textbf{Input}: $\cH\subseteq [0,1]^\cX$, $S=\lrset{\lr{x_i,y_i}}^m_{i=1}$.\\
    \textbf{Parameters}: Approximation parameter $\epsilon\in (0,1)$, weak learner sample size $d\geq 1$, sparsification parameter $k\geq 1$, number of boosting rounds $T\geq 1$, loss parameter $p\in \lrbra{1,\infty}$.\\
    \textbf{Algorithms used:} $\ell_p\text{-}\RERM_\cH$ (\cref{def:lp-rerm}), $\psi\text{-}\RERM_\cH$ (\cref{def:eta-rerm}), a variant of median boosting: $\mathrm{MedBoost}$ (\cref{alg:med-boost}), sparsification method (\cref{alg:Sparsify}).
    \begin{enumerate}
            \item Compute $h^\star\leftarrow \ell_p\text{-}\RERM_\cH(S)$. \\
            Denote $\psi(x,y)=\sup_{z\in\cU(x)}|h^\star(z)-y|$, $\forall (x,y)\in S$.
    	\item Inflate $S$ to $S_\cU$ to include all perturbed points. 
    	\item Discretize $\bar{S}_\cU\subseteq S_\cU$:
            (i) Construct a function class $\hat{\cH}$, where each $\hat{h}\in \hat{\cH}$ 
            is obtained by $\psi\text{-}\RERM$ optimizer operating on $d$ points from $S$. The input cutoff parameters to the optimizer are $\psi(x,y)$, as computed in step 1.\\
            (ii) Let $\Tilde{\cH}=\hat{\cH}\cup \lrset{h^\star}$. Each $\lr{z,y}\in S_\cU$ defines a function in the dual space, $f_{\lr{z,y}}: \Tilde{\cH} \rightarrow \lrbra{0,1}$ such that $f_{\lr{z,y}}(h) = \Lrabs{h(z)- y}^p$. 
         Define $\bar{S}_\cU$ to be the minimal cover of $S_\cU$ at scale $\cO \lr{\epsilon/p}$ under the supremum norm.
    	\item Compute robust $\medboost$ on $\bar{S}_\cU$, where $\hat{\cH}$ consists of weak learners for any distribution over $\bar{S_\cU}$.  Let $\cF=\lrset{\hat{h}_1,\ldots,\hat{h}_T}$ be the returned set of classifiers.
    	\item Sparsify the set $\cF$ to a smaller set $\lrset{\hat{h}_1,\ldots,\hat{h}_k}$.
    \end{enumerate}
    \textbf{Output:}
    $\hat{h}=\text{Median}\lr{\hat{h}_1,\ldots,\hat{h}_k}$.
\end{algorithm}

\paragraph{Proof overview and algorithm outline.}
The complete proof is in \cref{app:l1-regression}.

We explain the main differences from \cref{alg:lp-regression-highvote} and where the sample complexity improvement comes from.
In the discretization step, 
we replace the base learners in 
$\hat{\cH}$ with weak learners.
We construct an improper ensemble predictor via a median boosting algorithm, where the weak learners are chosen from $\hat{\cH}$. Specifically, each function in $\hat{\cH}$ is the output of $\psi\text{-}\RERM$ for $\cH$ (defined in \cref{def:eta-rerm}), performing on a subset $S'\subseteq S$ of size 
\[
\begin{dcases}
    d \leftarrow \Tilde{\cO}\lr{\fat\lr{\cH,c\epsilon/p}
    }, p\in [1,\infty)\\
   d\leftarrow
   \Tilde{\cO}\lr{\fat\lr{\cH,c\epsilon}
   }, p=\infty.
\end{dcases}
\]
This is in contrast to \cref{alg:lp-regression-highvote}, where we use $\mathrm{Multiplicative\;Weights}$ update that operates with stronger base learners. We can make accurate predictions by aggregating the outputs of the weak learners by taking their median instead of the average. 
Another improvement arises from sparsifying the ensemble \citep{hanneke2019sample} to be \emph{independent} of the sample size while keeping the median accurate almost with the same resolution. The sparsification step uses sampling and uniform convergence in the dual space (with respect to the non-robust loss).

We elaborate on the steps in \cref{alg:l1-regression}. Steps (1),(2), and (3) are similar to \cref{alg:lp-regression-highvote}, besides the construction of $\hat{\cH}$ as we explained above. In step (4), 
we compute a robust version of the real-valued boosting algorithm $\medboost$ \citep{kegl2003robust} on the discretized set $\bar{S}_\cU$, see \cref{alg:med-boost}. 
\citet{hanneke2019sample} showed how to construct a sample compression scheme from $\medboost$. From this step, we have that for any point in $\bar{S}_\cU$, the median of the losses of each hypothesis in the ensemble is within
$2\epsilon$ of the target cutoff that was computed in step 1. 
By the covering argument, the median of the losses is within $4\epsilon$ for any point in $(z,y)\in S_\cU$,
\begin{align*}
    \LRabs{\med\Lr{\hat{h}_1(z)-y,\ldots,\hat{h}_T(z)-y}}^p
    \leq
    \psi(z,y)^p + 4\epsilon.
\end{align*}
The median is translation invariant, so we have  
\begin{align*}
\LRabs{\med\Lr{\hat{h}_1(z),\ldots,\hat{h}_T(z)}-y}^p
    \leq
    \psi(z,y)^p + 4\epsilon.
\end{align*}
Finally, for any $(x,y)\in S$,
\begin{align*}
    \sup_{z\in\cU(x)}
    \LRabs{\med\Lr{\hat{h}_1(z)-y,\ldots,\hat{h}_T(z)-y}}^p
    \leq
    \psi(x,y)^p + 4\epsilon.
\end{align*}

To further reduce the sample compression size, 
in step (5) we sparsify the ensemble to $k=\Tilde{\cO}\lr{\fat^*\lr{\cH,c\epsilon}}$ functions,
\begin{align*}
    \sup_{z\in\cU(x)}
    \LRabs{\med\Lr{\hat{h}_1(z)-y,\ldots,\hat{h}_k(z)-y}}^p
    \leq
    \psi(x,y)^p + 5\epsilon.
\end{align*}
The proof follows by applying a sample compression generalization bound in the agnostic case, bounding the compression size, and rescaling $\epsilon$.

\section{Robust $(\eta,\beta)$-Regression}\label{sec:eta-beta-regression}

In this section, we study robust $(\eta,\beta)$-regression in realizable and agnostic settings. We provide an algorithm for the realizable setting and show how to reduce agnostic to realizable learning. We conclude by deriving sample complexity guarantees for both settings.

This model is different than regression which guarantees a small expected error (with high probability).
In the robust $(\eta,\beta)$-regression, we aim for a
small pointwise absolute error \emph{almost everywhere} on the support of the distribution. Results for this model do not follow from the standard regression model.
We first present our result for the realizable case. The proof is in \cref{app:eta-beta-regression}.
\begin{theorem}\label{thm:re-uni-reg}
Let $\cD$ be a distribution that is $\eta$-uniformly realizable (see \cref{def:eta-uni-re}) by a class $\cH \subseteq [0,1]^\cX$.
\cref{alg:improper-robust-learner-eta-beta} implies that
    the sample complexity for robust $(\eta,\beta,\epsilon,\delta)$-PAC learning a concept class $\cH$ is
        \begin{align*}
        \Tilde{\cO}\lr{\frac{\fat\lr{\cH,c\beta}{\fat^*\lr{\cH,c\beta}}}{\epsilon}+\frac{1}{\epsilon}\log\frac{1}{\delta}},
    \end{align*}
    for some numerical constant $c\in (0,\infty)$, 
    Recall that $\fat^*\lr{\cF,\epsilon} 
   \lesssim 
   \frac{1}{\epsilon}2^{\fat\lr{\cF,\epsilon/2}+1}$ by \cref{eq:dual-fat}.
\end{theorem}
\begin{algorithm}[H]
    \caption{Improper Robust $(\eta,\beta)$-Regressor for the Realizble Setting}\label{alg:improper-robust-learner-eta-beta}
    \textbf{Input}: $\cH\subseteq [0,1]^\cX$, $S=\lrset{\lr{x_i,y_i}}^m_{i=1}$.\\
    \textbf{Parameters}: Approximation parameters $\eta,\beta \in (0,1)$, sparsification parameter $k\geq 1$, number of boosting rounds $T\geq 1$.\\
    \textbf{Algorithms used:} $\psi\text{-}\RERM_\cH$ (\cref{def:eta-rerm}), a variant of median boosting: $\mathrm{MedBoost}$ (\cref{alg:med-boost}), sparsification method (\cref{alg:Sparsify}).
    \begin{enumerate}
    	\item Inflate $S$ to $S_\cU$ to include all perturbed points. 
            \item Discretize $\bar{S}_\cU\subseteq S_\cU$:
            (i) Construct a function class $\hat{\cH}$, where each $\hat{h}\in \hat{\cH}$ defined by $\psi\text{-}\RERM$ optimizer on $\Tilde{\cO}\lr{\fat\lr{\cH,\cO\lr{\beta}}}$ points from $S$. The input cutoff parameters to the optimizer are fixed $\eta$ for all points.\\
            (ii) Let $\Tilde{\cH}=\hat{\cH}\cup \lrset{h^\star}$. Each $\lr{z,y}\in S_\cU$ defines a function in the dual space, $f_{\lr{z,y}}: \Tilde{\cH} \rightarrow \lrbra{0,1}$ such that $f_{\lr{z,y}}(h) = \Lrabs{h(z)- y}$. 
            Define $\bar{S}_\cU$ to be the minimal cover of $S_\cU$ under $\lrnorm{\cdot}_\infty$ norm at scale $\cO\lr{\beta}$.
        \item Compute robust $\medboost$ on $\bar{S}_\cU$, where $\hat{\cH}$ consists of weak learners for any distribution over $\bar{S_\cU}$.  Let $\cF=\lrset{\hat{h}_1,\ldots,\hat{h}_T}$ be the returned set of classifiers.
    	\item Sparsify the set $\cF$ to a smaller set $\lrset{\hat{h}_1,\ldots,\hat{h}_k}$.
        \end{enumerate}
    \textbf{Output:}
    $\hat{h}=\text{Median}\lr{\hat{h}_1,\ldots,\hat{h}_k}$.
\end{algorithm}
We explain the main differences from \cref{alg:l1-regression}.
This model is different from robust regression with the $\ell_1$ loss. Our goal is to find a predictor with a prediction within $\eta+\beta$ of the true label \emph{almost everywhere} the domain, 
assuming that the distribution is $\eta$-uniformly realizable by the function class (\cref{def:eta-uni-re}). 

In this model, the cutoff parameter is given to us as a parameter and is \emph{fixed} for all points. This is different from \cref{alg:lp-regression-highvote,alg:l1-regression}, where we computed the changing cutoffs with a robust ERM oracle.
Moreover, the weak learners in $\hat{\cH}$ 
are defined as the output of $\psi\text{-}\RERM$ performing on a subset $S'\subseteq S$ of size $d=\Tilde{\cO}\lr{\fat\lr{\cH,\cO\lr{\beta}}}$. Note that the scale of shattering depends on $\beta$ and not $\epsilon$. The resolution of discretization in the cover depends on $\beta$ as well.
\paragraph{Agnostic setting}
We establish an upper bound on the sample complexity of the agnostic setting, by using a reduction to the realizable case. The main argument was originally suggested in \cite{david2016supervised} for the 0-1 loss and holds for the $\eta$-ball robust loss as well. The proof is in \cref{app:eta-beta-regression}.

\begin{theorem}\label{thm:ag-uni-reg}
    The sample complexity for agnostic robust $(\eta,\beta,\epsilon,\delta)$-PAC learning a concept class $\cH$ is
        \begin{align*}
        \Tilde{\cO}\lr{\frac{\fat\lr{\cH,c\beta}{\fat^*\lr{\cH,c\beta}}}{\epsilon^2}+\frac{1}{\epsilon^2}\log\frac{1}{\delta}},
    \end{align*}
    for some numerical constant $c\in (0,\infty)$.

    Recall that $\fat^*\lr{\cF,\epsilon} 
   \lesssim 
   \frac{1}{\epsilon}2^{\fat\lr{\cF,\epsilon/2}+1}$ by \cref{eq:dual-fat}.
\end{theorem}
\begin{remark}
    An agnostic learner for robust $(\eta,\beta)$-regression does not apply to the robust regression setting. The reason is that the optimal function in $\cH$ may have different scales of robustness on different points, which motivates our approach of using changing cutoffs for different points.
In \cref{app:naive-approach-fixed-cutoff} we show that by using a fixed cutoff for all points we can obtain an error of only $\sqrt{\text{OPT}_\cH}+\epsilon$.
\end{remark} 

\section{Discussion}
In this paper, we studied the robustness of real-valued functions to test time attacks. 
We showed that finite fat-shattering is sufficient for learnability. 
we proved sample complexity for learning with the general $\ell_p$ losses and improved it for the $\ell_1$ loss. We also studied a model of regression with a cutoff loss. We proved sample complexity in realizable and agnostic settings.
We leave several interesting open questions for future research.
(i) Improve the upper bound for learning with $\ell_p$ robust loss (if possible) and show a lower bound. There might be a gap between sample complexities of different values of $p$. More specifically, what is the sample complexity for learning with $\ell_2$ robust loss?
(ii) We showed that the fat-shattering dimension is a sufficient condition. What is a necessary condition? In the binary-valued case, we know that having a finite $\VC$ is not necessary.
(iii) To what extent can we benefit from unlabeled samples for learning real-valued functions? This question was considered by \citet{attias2022characterization} for binary function classes, where they showed that the labeled sample complexity can be arbitrarily smaller compared to the fully-supervised setting.
 (iv) In this work we focused on the statistical aspect of robustly learning real-valued functions. It would be interesting to explore the computational aspect as well.

\section*{Acknowledgments}
We thank Meni Sadigurschi for helpful discussions regarding sample compression schemes of real-valued functions.
We are grateful to Uri Sherman for helpful discussions on the generalization of ERM in Stochastic Convex Optimization.
We would like to thank Arvind Ramaswami for pointing out a mistake in the sample compression definition in our previous version (the change does not affect the correctness of the proofs).

This project has received funding from the European Research Council (ERC) under the European Union’s Horizon 2020 research and innovation program (grant agreement No. 882396), by the Israel Science Foundation (grants 993/17, 1602/19), Tel Aviv University Center for AI and Data Science (TAD), and the Yandex Initiative for Machine Learning at Tel Aviv University.
I.A. is supported by the Vatat Scholarship from the Israeli Council for Higher Education and by the Kreitman School of Advanced Graduate Studies.

\bibliography{paperbib}

\appendix

\section{Auxiliary Results}\label{app:auxiliary-section}

\begin{theorem}[Generalization from approximate interpolation]\citep[Theorems 21.13 and 21.14]{anthony1999neural}\label{thm:generalization-interpolation}
Let $\cF\subseteq [0,1]^\cX$ be a function class with a finite fat-shattering dimension (at any scale). For any $\eta,\beta,\epsilon,\delta\in(0,1)$, any distribution $\cD$ over $\cX$, any function $t:\cX \rightarrow [0,1]$, for a random sample $S\sim \cD^m$, if 
$$
m(\eta, \beta,\epsilon,\delta)=\cO\lr{\frac{1}{\epsilon}\lr{\fat\lr{\cF,\beta/8}\log^2\lr{\frac{\fat\lr{\cF,\beta/8}}{\beta\epsilon}}+\log\frac{1}{\delta}}},
$$ 
then with probability at least $1-\delta$ over $S$, for any ${f}\in\cF$ satisfying $\lrabs{{f}(x)-t(x)}\leq \eta$ $\forall{(x,y)\in S}$,
it holds that $\hP_{x\sim\cD}\lrset{x: \lrabs{f(x)-t(x)} \leq \eta+\beta}\geq 1-\epsilon$.
\end{theorem}

The following is a bound on the covering numbers in $\lrnorm{\cdot}_\infty$.
\begin{lemma}[\textbf{Covering numbers for infinity metric}]\citep[Theorem 4.4]{rudelson2006combinatorics}\label{lem:covering-num}
Let $\cF\subseteq [0,1]^\Omega$ be a class of functions and $\lrabs{\Omega}=n$. Then for any $0<a\leq 1$ and $0<t<1/2$,
\begin{align*}
    \log\cN\lr{t,\cF,\lrnorm{\cdot}_\infty}\leq Cv\log\lr{n/vt}\cdot\log^a\lr{2n/v},
\end{align*}
where $v=\fat_{ca t}(\cF)$, and $C,c$ are universal constants.
\end{lemma}
The following generalization for sample compression in the realizable case was proven by \citet{littlestone1986relating,floyd1995sample}. Their proof is for the 0-1 loss, but it applies similarly to bounded loss functions.
\begin{lemma}[\textbf{Sample compression generalization bound}]\label{lem:sample-compression}
Let a sample compression scheme $(\kappa,\rho)$, and a loss function $\ell:\hR\times\hR\rightarrow [0,1]$. In the realizable case, for any $\kappa(S)\lesssim m$, any $\delta\in (0,1)$, and any distribution $\cD$ over $\cX\times\lrset{0,1}$, for $S\sim \cD^m$, with probability $1-\delta$, 
{
\begin{align*}
{\err}\lr{\rho({\kappa({S})});\cD} \leq
\widehat{\err}\lr{\rho({\kappa({S})});S}
+
\cO\lr{\frac{|\kappa(S)|\log(m)+\log \frac{1}{\delta}}{m}}.
\end{align*}
}%
\end{lemma}
The following generalization for sample compression in the agnostic case was proven by \citet{graepel2005pac}. Their proof is for the 0-1 loss, but it applies similarly to bounded loss functions. We use it with the $\eta$-ball robust loss.
\begin{lemma}[\textbf{Agnostic sample compression generalization bound}]
\label{lem:sample-compression-agnostic}
Let a sample compression scheme $(\kappa,\rho)$, and a loss function $\ell:\hR\times\hR\rightarrow [0,1]$. In the agnostic case, for any $\kappa(S)\lesssim m$, any $\delta\in (0,1)$, and any distribution $\cD$ over $\cX\times\lrset{0,1}$, for $S\sim \cD^m$, with probability $1-\delta$, 
{
\begin{align*}
{\err}\lr{\rho({\kappa({S})});\cD}
\leq
\widehat{\err}\lr{\rho({\kappa({S})});S}
+
\cO\lr{\sqrt{\frac{|\kappa(S)|\log(m)+\log \frac{1}{\delta}}{m}}}.
\end{align*}
}%
\end{lemma}

\section{Proof of \cref{thm:generalization-interpolation-cutoffs}: Generalization from Approximate Interpolation with Changing Cutoffs}
\label{app:generalization-interpolation}
 Let $\cF\subseteq [0,1]^{\cX}$ and let 
$$
\cH=\lrset{(x,y)\mapsto\lrabs{f(x)-y}:f\in\cF}.
$$
Define the function classes 
$$\cF_1= \lrset{(x,y)\mapsto \lrabs{f(x)-y}-\psi(x,y):f\in \cF},$$
and 
$$\cF_2= \lrset{(x,y)\mapsto \max\lrset{f(x,y),0}:f \in \cF_1}.$$

We claim that $\fat(\cH,\gamma)=\fat(\cF_1,\gamma)$. Take a set $S=\lrset{(x_1,y_1),\ldots,(x_m,y_m)}$ that is $\gamma$-shattered by $\cH$. 
There exists a witness $r = (r_1, \ldots,r_m) \in [0,1]^m$ such that for each $\sigma = (\sigma_1, \ldots, \sigma_m) \in \{-1, 1\}^m$ there is a function $h_{\sigma} \in \cH$ such that
\[
\forall i \in [m] \; 
\begin{cases}
	h_{\sigma}((x_i,y_i)) \geq r_i + \gamma, & \text{if $\sigma_i = 1$}\\
    h_{\sigma}((x_i,y_i)) \leq r_i - \gamma, & \text{if $\sigma_i = -1$.}
 \end{cases}
\]
The set $S$ is shattered by $\cF_1$
by taking $\Tilde{r}= \lr{r_1+\psi(x_1,y_1),\ldots,r_m+\psi(x_m,y_m)}$. Similarly, any set that is shattered by $\cF_1$ is also shattered by $\cH$.

The class $\cF_2$ consists of choosing a function from $\cF_1$ and computing its pointwise maximum with the constant function 0.
In general, for two function classes $\cG_1,\cG_2$, we can define the maximum aggregation class
\begin{equation*}
 \max(\cG_1,\cG_2)=\{x\mapsto\max\lrset{g_1(x),g_2(x)}:g_i \in \cG_i\},
\end{equation*}
\citet{kontorovich2021fat} showed that for any $\cG_1,\cG_2$
\begin{align*}
 \fat\lr{ \max(\cG_1,\cG_2),\gamma}   \lesssim
\lr{\fat\lr{\cG_1,\gamma}+\fat\lr{\cG_2,\gamma}}\log^2\lr{\fat\lr{\cG_1,\gamma}+\fat\lr{\cG_2,\gamma}}.
\end{align*}
Taking $\cG_1=\cF_1$ and $\cG_2\equiv 0$, we get 
\begin{align*}
 \fat\lr{\cF_2,\gamma}   \lesssim
\fat\lr{\cF_1,\gamma}\log^2\lr{\fat\lr{\cF_1,\gamma}}.
\end{align*}
For the particular case $\cG_2\equiv 0$, we can show a better bound of 
\begin{align*}
 \fat\lr{\cF_2,\gamma}   \lesssim
\fat\lr{\cF_1,\gamma}.
\end{align*}
In words, it means that truncation cannot increase the shattering dimension.
Indeed, take a set $S=\lrset{(x_1,y_1),\ldots,(x_k,y_k)}$ that is $\gamma$-shattered by $\cF_2=\max\lr{\cF_1,0}$, we show that this set is  $\gamma$-shattered by $\cF_1$.
There exists a witness $r = (r_1, \ldots,r_k) \in [0,1]^k$ such that for each $\sigma = (\sigma_1, \ldots, \sigma_k) \in \{-1, 1\}^k$ there is a function $f_{\sigma} \in \cF_1$ such that
\[
\forall i \in [k] \; 
\begin{cases}
    \max\lrset{f_{\sigma}((x_i,y_i)),0} \geq r_i + \gamma, & \text{if $\sigma_i = 1$}\\
    \max\lrset{f_{\sigma}((x_i,y_i)),0} \leq r_i - \gamma, & \text{if $\sigma_i = -1$.}
 \end{cases}
\]
For $\max\lrset{f_{\sigma}((x_i,y_i)),0} \leq r_i - \gamma$, we simply have that $f_{\sigma}((x_i,y_i)) \leq r_i - \gamma$. Moreover, this implies that $r_i\geq \gamma$. As a result, 
\begin{align*}
     \max\lrset{f_{\sigma}((x_i,y_i)),0} 
     &\geq 
     r_i + \gamma
     \\
     &\geq
     2\gamma
    \\
     &>
     0,
\end{align*}
which means that $f_{\sigma}((x_i,y_i)) \geq r_i + \gamma$. This shows that $\cF_1$ $\gamma$-shatters $S$ as well.
We can conclude the proof by applying \cref{thm:generalization-interpolation} to the class $\cF_2$ with $t(x)=0$ and $\eta=\beta$.

\section{Proofs for \cref{sec:lp-regression}: Robust Regression for $\ell_p$ Losses}\label{app:lp-regression}

\begin{proof}[of \cref{thm:lp-regression}]
Fix $\epsilon,\delta \in (0,1)$ and $p\in [1,\infty]$.
Let $\cH \subseteq [0,1]^\cX$. Fix a distribution $\cD$ over $\cX\times \cY$, and let $S=\lrset{\lr{x_i,y_i}}^m_{i=1}$ be an i.i.d. sample from $\cD$. We first prove for $p\in \lrset{1,\infty}$, and generalize for $p\in (1,\infty)$ by using the Lipschitzness of the $\ell_p$ loss. 
We follow the steps as described in 
 \cref{alg:lp-regression-highvote}.

\begin{enumerate}[leftmargin=0.5cm]
    \item Compute ${h^\star}\leftarrow \ell_p\text{-}\RERM_\cH(S)$ in order to get the set of cutoffs $\psi(x,y)=\sup_{z\in\cU(x)}|h^\star(z)-y|$ for $(x,y)\in S$. Let $\psi|_S=\lr{\psi(x_1,y_1),\ldots,\psi(x_m,y_m)}$.
    Our goal is to construct a predictor with an empirical robust loss of $\psi(x,y)^p+\epsilon$, for $p\in (1,\infty)$, and $\psi(x,y)+\epsilon$ for $p\in \lrset{1,\infty}$, for any $(x,y)\in S$, which means that our predictor is an approximate robust ERM.
    \item Define the inflated training data set 
    $$S_\cU = \bigcup_{i
    \in [n]}\lrset{(z,y_{I(z)}):z\in \cU(x_i)},$$ 
    where $I(z) =\min\lrset{i\in [m]:z\in \cU(x_i)}$. 
    For $(z,y)\in S_\cU$, let $\psi(z,y)$ be the $\psi(x,y)$ for which $z \in \cU(x)$ and $y_{I(z)}=y$.
    \item Discretize $S_{\cU}$ to a finite set $\bar{S}_\cU$ as follows. 
    \begin{enumerate}[leftmargin=0.6cm]
        \item Define a set of functions, such that each function
        is defined by an $\epsilon$-approximate $\psi\text{-}\RERM_{\cH}$ optimizer on 
        $d=\cO\lr{\frac{1}{\epsilon}\fat\lr{\cH,\epsilon/8}\log^2\lr{\frac{\fat\lr{\cH,\epsilon/8}}{\epsilon^2}}}$ points from $S$,
        
    $$\hat{\cH}=\lrset{\psi\text{-}\RERM_{\cH}(S',\psi|_{S'},\epsilon): S'\subseteq S, |S'|=d}.$$
    Recall the definition of $\psi\text{-}\RERM_\cH$, see \cref{def:eta-rerm}. 
    The cardinality of this class is bounded as follows
    \begin{align}\label{eq:H-hat}
    |\hat{\cH}|
    \approx
    {m \choose d}
    \lesssim
    \left(\frac{m}{d}\right)^{d}.    
    \end{align}
      
 \item A discretization $\bar{S}_\cU\subseteq S_\cU$ will be defined by covering of the dual class in $\lrnorm{\cdot}_\infty$. Define $\Tilde{\cH}=\hat{\cH}\cup \lrset{h^\star}$.
    Let $L^1_{\Tilde{\cH}}$ be the $L_1$ loss class of $\Tilde{\cH}$, namely, $L^1_{\Tilde{\cH}}=\lrset{\cZ\times\cY \ni (z,y)\mapsto |h(z)-y|: h \in \Tilde{\cH}}$.
    The \textit{dual class} of $L^1_{\Tilde{\cH}}$ ,
    ${L^1_{\Tilde{\cH}}}^* \subseteq \lrbra{0,1}^{\Tilde{\cH}}$,
    is defined as the set of all functions $f_{\lr{z,y}}: \Tilde{\cH} \rightarrow \lrbra{0,1}$ such that $f_{\lr{z,y}}(h) = \Lrabs{h(z)- y}$,
    for any $(z,y)\in S_{\cU}$. 
    Formally, ${L^1_{\Tilde{\cH}}}^*=\lrset{f_{\lr{z,y}}:(z,y)\in S_\cU}$, where $f_{\lr{z,y}}= \lr{f_{(z,y)}(h_1),\ldots,f_{(z,y)}(h_{\lrabs{\Tilde{\cH}}})}$.
    We take $\bar{S}_\cU \subseteq S_\cU$ to be a minimal $\epsilon$-cover for $S_\cU$ in $\lrnorm{\cdot}_\infty$,
    \begin{align}\label{eq:covering-lp}
    \sup_{\lr{z,y}\in S_\cU}\inf_{\lr{\bar{z},\bar{y}}\in \bar{S}_\cU} \lrnorm{f_{\lr{z,y}}-f_{\lr{\bar{z},\bar{y}}}}_\infty\leq \epsilon.  
    \end{align}
    Let $\fat^*\lr{L^1_{\Tilde{\cH}},\epsilon}$ be the dual $\epsilon$-fat-shattering of $L^1_{\Tilde{\cH}}$.
    Applying a covering number argument from \cref{lem:covering-num} on the dual space and upper bounding the dual fat-shattering of the $L_1$ loss class with the dual fat-shattering of $\Tilde{H}$,
    we have the following bound
    \begin{align}\label{eq:covering-l1}
    \begin{split}
    \lrabs{\bar{S}_\cU}
    &=
    \cN\lr{\epsilon,S_\cU,\lrnorm{\cdot}_\infty}
    \\
    &\lesssim 
    \mathrm{exp}\lr{{ \fat^*\lr{L^1_{\Tilde{\cH}},c\epsilon}\log^2\lr{\frac{|\Tilde{\cH}|}{\epsilon}}}}
    \\
    &\lesssim 
    \mathrm{exp}\lr{{ \fat^*\lr{\Tilde{\cH},c\epsilon}\log^2\lr{\frac{|\Tilde{\cH}|}{\epsilon}}}}
    \\
    &\lesssim 
    \mathrm{exp}\lr{{ \fat^*\lr{\cH,c\epsilon}\log^2\lr{\frac{|\Tilde{\cH}|}{\epsilon}}}},
    \end{split}
    \end{align}
    \end{enumerate}
 where $c\in(0,\infty)$ is a numerical constant, derived from the covering argument in \cref{lem:covering-num}.  

 \item Compute the following robust variant of $\mathrm{Multiplicative\;Wights\;(MW)}$ algorithm on the discretized set $\bar{S}_\cU$ for $T\approx\log\lrabs{\bar{S}_\cU}$.
 Let $d=\cO\lr{\frac{1}{\epsilon}\fat\lr{\cH,\epsilon/8}\log^2\lr{\frac{\fat\lr{\cH,\epsilon/8}}{\epsilon^2}}}$, and let $\psi(\bar{z},\bar{y})$ be the $\psi(x,y)$ for which $\bar{z} \in \cU(x)$.

From \cref{thm:generalization-interpolation-cutoffs}, taking $\delta=1/3$, $\beta=\epsilon$, we know that for any distribution $\cP$ on $\bar{S}_\cU$, upon receiving an i.i.d. sample $S''$ from $\cP$ of size $d$, with probability $2/3$ over sampling $S''$ from $\cP$, for any ${h}\in\cH$ with 
$\forall (\bar{z},\bar{y})\in S'':\; \lrabs{{h}(\bar{z})-\bar{y}}\leq \psi(\bar{z},\bar{y})+\epsilon$, 
it holds that \\
$\hP_{(\bar{z},\bar{y})\sim\cP}\lrset{(\bar{z},\bar{y}): \lrabs{h(\bar{z})-\bar{y}}\leq\psi(\bar{z},\bar{y})+2\epsilon}\geq 1-\epsilon$. We can conclude that for any distribution $\cP$ on $\bar{S}_\cU$, there exists such a set of points $S''\subseteq \bar{S}_\cU$.

Given that set, we can find the function with the aforementioned property in $\hat{\cH}$.
Let $S'$ be the $d$ points in $S$ that the perturbed points $S''$ originated from.
That is, $S''\subseteq \bigcup_{(x,y)\in {S'}}\bigcup \lrset{(\bar{z},\bar{y}): \bar{z}\in \cU(x)}$.
Take $\hat{\cH}\ni\hat{h} = \psi\text{-}\RERM_\cH(S',\psi|_{S'},\epsilon$), it holds that 
$\forall (\bar{z},\bar{y})\in S'':\; \lrabs{{\hat{h}}(\bar{z})-\bar{y}}\leq \psi(\bar{z},\bar{y})+\epsilon$, 
as a result we get $\hP_{(\bar{z},\bar{y})\sim\cP}\lrset{(\bar{z},\bar{y}): \lrabs{\hat{h}(\bar{z})-\bar{y}}\leq\psi(\bar{z},\bar{y})+2\epsilon}\geq 1-\epsilon$. 
\begin{algorithm}[H]
\caption{Robust $\mathrm{Multiplicative\; Weights}$}\label{alg:MW}
\textbf{Input:} $\cH, S, \bar{S}_\cU$.
\\
\textbf{Parameters:} Approximation parameter $\epsilon\in (0,1)$, weights update parameter $\xi\in (0,1)$, number of boosting rounds $T\geq1$, base learner sample size $d\geq 1$, loss parameter $p\in [1,\infty]$ cutoff parameters $\psi|_S=\lr{\psi(x_1,y_1),\ldots,\psi(x_m,y_m)}$ for $(x_i,y_i),
\in S$ and $\psi(\bar{z},\bar{y})$ is the $\psi(x,y)$ for which $\bar{z} \in \cU(x)$.\\
\textbf{Algorithms used:} $\epsilon$-approximate $\psi$-robust empirical risk minimizer $\psi\text{-}\RERM_\cH$ (\cref{def:eta-rerm}).\\
\textbf{Initialize} $P_1$ = Uniform($\bar{S}_\cU$).\\
 For $t=1,\ldots,T$:
    \begin{enumerate}
        \item[]\textcolor{DarkBlue}{$\triangleright$ Compute a strong base learner w.r.t. distribution $\cP_t$ by finding $n$ points in $S$ and executing $\psi\text{-}\RERM_\cH$ on them.}
            \item 
            Find $d$ points ${S''_t}\subseteq \bar{S}_\cU$ such that any $h\in \cH$ satisfying:
            $\forall \lr{\bar{z},\bar{y}}\in S''_t:\; \lrabs{{h}(\bar{z})-\bar{y}}^p\leq \psi(\bar{z},\bar{y})^p+\epsilon$, 
            it holds that 
            $\hE_{(\bar{z},\bar{y})\sim\cP_t} \LRbra{\hI \LRset{\lrabs{h(\bar{z})- \bar{y}}^p \leq \psi(\bar{z},\bar{y})^p+2\epsilon}}\geq 1-\epsilon$.
            (See the analysis for why this set exists).

            \item Let $S'_t$ be the $d$ points in $S$ that 
            $S''_t$ originated from. Formally, $S''_t\subseteq \bigcup_{(x,y)\in {S'_t}}\bigcup \lrset{(\bar{z},\bar{y}): \bar{z}\in \cU(x)}$.
            \item Compute $\hat{h}_t = \psi\text{-}\RERM_\cH(S'_t,\psi|_{S'_t},\epsilon)$.
            \item[]\textcolor{DarkBlue}{$\triangleright$ Make a multiplicative weights update on $\cP_t$.}
        \item For each $(\bar{z},\bar{y})\in \bar{S}_\cU$:
        \begin{itemize}
            \item[] $P_{t+1}(\bar{z},\bar{y}) \propto P_{t}(\bar{z},\bar{y})e^{-\xi \hI\lrset{\lrabs{\hat{h}_t(\bar{z})-\bar{y}}^p\leq \psi(\bar{z},\bar{y})^p+2\epsilon}}$
        \end{itemize}
    \end{enumerate}
\textbf{Output:} classifiers $\hat{h}_1,\ldots,\hat{h}_T$ and sets $S'_1,\ldots,S'_T$.
\end{algorithm}

\end{enumerate}
\noindent\textbf{A uniformly $5\epsilon$-approximate adversarially robust sample compression scheme for $S$.}
The output of the algorithm is a sequence of functions $\hat{h}_1,\ldots,\hat{h}_T$, and the corresponding sets that encode them $S'_1,\ldots,S'_T$, where we predict with the average of the returned hypotheses, $\frac{1}{T}\sum^T_{t=1}\hat{h}_t(\cdot)$.
For $T\approx \log\lrabs{\bar{S}_\cU}$, we show that 
\begin{align}\label{eq:Su-bar-average}
\forall \lr{\bar{z},\bar{y}}\in \bar{S}_\cU:
\;\frac{1}{T}\sum^T_{t=1}\lrabs{\hat{h}_t(\bar{z})-\bar{y}}
\leq
\psi\lr{\bar{z},\bar{y}}+3\epsilon.
\end{align}
For any distribution $\cP_t$ over $\bar{S}_\cU$, we have a base learner $\hat{h}_t$, satisfying 
$\hE_{(\bar{z},\bar{y})\sim\cP_t} \LRbra{\hI \LRset{\lrabs{\hat{h}_t(\bar{z})- \bar{y}} \leq \psi(\bar{z},\bar{y})+2\epsilon}}\geq 1-\epsilon$, due to \cref{thm:generalization-interpolation-cutoffs}. Following standard analysis of $\mathrm{MW}$ / $\alpha$-$\mathrm{Boost}$ (see \citet[Section 6]{schapire2013boosting}), for any $\lr{\bar{z},\bar{y}}\in \bar{S}_\cU$, $1-\epsilon$ fraction of the base learners have an error within $\psi\lr{\bar{z},\bar{y}}+2\epsilon$. The loss is bounded by $1$, so the other $\epsilon$ fraction can add an error of at most $\epsilon$. The overall average loss of the base learners is upper bounded by $\psi\lr{\bar{z},\bar{y}}+3\epsilon$.
Note that we can find these base learners in $\hat{\cH}$, as defined in step $2(a)$ of the main algorithm.
Crucially, we use strong base learners in order to ensure a low empirical loss of the average base learners. 

From the covering argument (\cref{eq:covering-lp}), we have
\begin{align}\label{eq:Su-lp}
\forall \lr{z,y}\in S_\cU:\;
\frac{1}{T}\sum^T_{t=1}\lrabs{\hat{h}_t(z)-y}
\leq
\psi\lr{z,y}+5\epsilon.
\end{align}

Indeed, for any $\lr{z,y}\in S_\cU$ there exists $\lr{\bar{z},\bar{y}}\in \bar{S}_\cU$, such that for any 
$h\in \Tilde{\cH}$,
\begin{align*}
\LRabs{\Lrabs{h(z)-y}-\Lrabs{h(\bar{z})-\bar{y}}}\leq \epsilon.
\end{align*}
Specifically, it holds for 
$\lrset{\hat{h}_1,\ldots,\hat{h}_T}\subseteq \Tilde{\cH}$ and $h^\star\in\Tilde{\cH}$, and so
\begin{align}\label{eq:triangle-average}
 \frac{1}{T}\sum^T_{t=1}\lrabs{\hat{h}_t(z)-y}
\leq
\frac{1}{T}\sum^T_{t=1}\lrabs{\hat{h}_t(\bar{z})-\bar{y}} + \epsilon,   
\end{align}
and 
\begin{align}\label{eq:triangle-h-star}
  \psi(\bar{z},\bar{y})
=
\Lrabs{h^\star(\bar{z})-\bar{y}}
\leq
\Lrabs{h^\star(z)-y}+\epsilon
= 
\psi(z,y)+\epsilon.  
\end{align}
Combining \cref{eq:triangle-average,eq:triangle-h-star} we get \cref{eq:Su-lp}.
By using the convexity of the $\ell_1$ loss, we have 
\begin{align}\label{eq:convexity}
    \lrabs{\frac{1}{T}\sum^T_{t=1}\hat{h}_t(z)-y}
    \leq
    \frac{1}{T}\sum^T_{t=1}\lrabs{\hat{h}_t(z)-y}.
\end{align}

Finally, from \cref{eq:Su-lp,eq:convexity} we conclude a uniformly $5\epsilon$-approximate \emph{adversarially robust} sample compression scheme for $S$,
\begin{align}\label{eq:adv-compression-lp}
    \forall \lr{z,y}\in S_\cU:\;
    \lrabs{\frac{1}{T}\sum^T_{t=1}\hat{h}_t(z)-y}
    \leq
    \psi(z,y)+5\epsilon,
\end{align}
which implies that 
\begin{align*}
    \forall \lr{x,y}\in S:\;
    \sup_{z\in\cU(x)}\lrabs{\frac{1}{T}\sum^T_{t=1}\hat{h}_t(x)-y}
    \leq
    \psi(x,y)+5\epsilon.
\end{align*}

\paragraph{Bounding the compression size.} 
We have $T=\cO\lr{\log\lrabs{\bar{S}_\cU}}$ hypotheses, where each one is representable by $d=\cO\lr{\frac{1}{\epsilon}\fat\lr{\cH,\epsilon/8}\log^2\lr{\frac{\fat\lr{\cH,\epsilon/8}}{\epsilon^2}}}$ points. By counting the number of predictors using \cref{eq:covering-l1}, we get for $m\geq 2d$
\begin{align*}
    \begin{split}
    k
    &=
    \log\lr{\lrabs{\bar{S}_\cU}}
    \\
    &\lesssim
    \fat^*\lr{\cH,c\epsilon}\log^2\lr{\frac{|\Tilde{\cH}|}{\epsilon}}
    \\
    &\lesssim 
    \fat^*\lr{\cH,c\epsilon}\log^2\lr{\frac{1}{\epsilon}\lr{\lr{\frac{m}{d}}^d+1}}
    \\
    &\lesssim 
    \fat^*\lr{\cH,c\epsilon}\log^2\lr{\frac{1}{\epsilon}\lr{\frac{m}{d}}^d}
        \\
    &\lesssim 
    \fat^*\lr{\cH,c\epsilon}\lr{\log\lr{\frac{1}{\epsilon}}+d\log\lr{\frac{m}{d}}}^2
    \\
    &\lesssim 
    \fat^*\lr{\cH,c\epsilon}\lr{\log^2\lr{\frac{1}{\epsilon}}+\log\lr{\frac{1}{\epsilon}}d\log\lr{\frac{m}{d}}+d^2\log^2\lr{\frac{m}{d}}}
    \\
    &\lesssim 
    \fat^*\lr{\cH,c\epsilon}\log^2\lr{\frac{1}{\epsilon}}d^2\log^2\lr{\frac{m}{d}}.
    \end{split}
    \end{align*}
We get a uniformly $5\epsilon$-approximate \emph{adversarially robust} sample compression scheme for $S$ of size 
\begin{align*}
   \cO
\lr{
\fat^*\lr{\cH,c\epsilon}\log^2\lr{\frac{1}{\epsilon}}d^3\log^2\lr{\frac{m}{d}}
     }
     . 
\end{align*}

Each weak learner is encoded by a multiset $S'\subseteq S$ of size $d$ and is constructed by computing some $\hat{h}\in \cH$ that solves the  constrained optimization 
$$ \sup_{z\in\cU(x)}\lrabs{\hat{h}(z)-y}\leq \psi(x,y)+\epsilon, \; \forall (x,y)\in S'.$$
By plugging in $d=\cO\lr{\frac{1}{\epsilon}\fat\lr{\cH,\epsilon/8}\log^2\lr{\frac{\fat\lr{\cH,\epsilon/8}}{\epsilon^2}}}$, we have
\begin{align*}
\cO
\lr{
\frac{1}{\epsilon^3}\fat^3\lr{\cH,\epsilon/8}\fat^*\lr{\cH,c\epsilon}
\log^6\lr{\frac{\fat\lr{\cH,\epsilon/8}}{\epsilon^2}}
\log^2\lr{\frac{1}{\epsilon}}\log^2\lr{\frac{m}{\frac{1}{\epsilon}\fat\lr{\cH,\epsilon/8}\log^2\lr{\frac{\fat\lr{\cH,\epsilon/8}}{\epsilon^2}}}}}.
\end{align*}

\paragraph{Encoding base learners.} We encode each $\psi(x,y)$ by some approximation $\Tilde{\psi}(x,y)$, such that \\$\lrabs{\Tilde{\psi}(x,y)-\psi(x,y)}\leq \epsilon$, by discretizing $[0,1]$ to $1/\epsilon$ buckets of size $\epsilon$, and each $\psi(x,y)$ is rounded down to the closest value $\Tilde{\psi}(x,y)$. Each approximation requires to encode $\log\lr{1/\epsilon}$ bits, and so each learner encodes $d\log\lr{1/\epsilon}$ bits and $d$ samples. We have $k$ weak learners, and the compression size is 
\begin{align*}
    k(d+d\log\lr{1/\epsilon})
    &\leq
    2kd\log\lr{1/\epsilon}.
\end{align*}
Therefore, we have a uniform $5\epsilon$-approximate compression for $\ell_1$ and $\ell_\infty$ losses of size
\begin{align*}
\cO
\lr{
\frac{1}{\epsilon^3}\fat^3\lr{\cH,\epsilon/8}\fat^*\lr{\cH,c\epsilon}
\log^6\lr{\frac{\fat\lr{\cH,\epsilon/8}}{\epsilon^2}}
\log^3\lr{\frac{1}{\epsilon}}\log^2\lr{\frac{m}{\frac{1}{\epsilon}\fat\lr{\cH,\epsilon/8}\log^2\lr{\frac{\fat\lr{\cH,\epsilon/8}}{\epsilon^2}}}}}.
\end{align*}
\paragraph{Generalizing for $p\in(1,\infty)$.} 
We rely on the Lipschitzness of the $\ell_p$ loss and rescaling the approximation parameter $\epsilon$ to $\epsilon/p$.
Recall the covering of $S_\cU$ in step 3(b).
Note that an $(\epsilon/p)$-cover for the $L_1$ loss class is an $\epsilon$-cover for the $L_p$ loss class due to the Lipschitzness of the $\ell_p$ loss
    \begin{align*}
        \LRabs{\lrabs{h(z)-y}^p-\lrabs{h(\bar{z})-\bar{y}}^p}
        &\leq
        p\LRabs{\lrabs{h(z)-y}-\lrabs{h(\bar{z})-\bar{y}}}
        \\
        &\leq
        p\epsilon.
    \end{align*}

Moreover, we constructed a function $f=\frac{1}{T}\sum^T_{t=1}\hat{h}_t(x)$ with $\sup_{z\in\cU(x)}\lrabs{f(z)-y} \leq \psi(x,y)+\epsilon$ for any $(x,y)\in S$. Note that since $\lrabs{f(\cdot)-y}\in [0,1]$, the same $z$ that maximizes the $\ell_1$ loss also maximizes for any $\ell_p$.
This implies that 
\begin{align*}
\sup_{z\in\cU(x)}\lrabs{f(z)-y}^p 
\overset{(i)}{\leq} 
\lr{\lr{\psi(x,y)}+\epsilon}^p 
\overset{(ii)}{\leq}
\psi(x,y)^p+p\epsilon,
\end{align*}
where (i) follows by just raising both sides to the power of $p$ and
(ii) follows since the function $x\mapsto \lrabs{x-y}^p$ is $p$-Lipschitz for $(x-y)\in [0,1]$, and so
\begin{align*}
    \lrabs{\lr{\psi(x,y)+\epsilon}^p - \psi(x,y)^p}
    &\leq
    p\lrabs{\psi(x,y)+\epsilon - \psi(x,y)}
    \\
    &\leq
    p\epsilon.
\end{align*}
By rescaling $p\epsilon$ to $\epsilon$, we get 
\begin{align*}
        \sup_{z\in\cU(x)}\lrabs{\frac{1}{T}\sum^T_{t=1}\hat{h}_t(x)-y}^p 
        &\leq 
        \psi(x,y)^p+\epsilon.
\end{align*}
%
Therefore, we have an approximate compression for $\ell_p$ of size
\begin{align*}
\cO
\lr{
\frac{1}{\epsilon^3}\fat^3\lr{\cH,c\epsilon/p}\fat^*\lr{\cH,c\epsilon/p}
\log^6\lr{\frac{p^2\,\fat\lr{\cH,\epsilon/8}}{\epsilon^2}}
\log^3\lr{\frac{p}{\epsilon}}\log^2\lr{\frac{m}{\frac{1}{\epsilon}\fat\lr{\cH,c\epsilon/p}\log^2\lr{\frac{p^2\,\fat\lr{\cH,c\epsilon/p}}{\epsilon^2}}}}}.
\end{align*}
Note that the $1/\epsilon^3$ term is not affected by the rescaling. The instances of $\epsilon$ that should be rescaled are the ones that arise from the covering argument in step 3(b), and the approximation parameter for the sample compression.
\paragraph{Generalization bound.}
Let $(\kappa,\rho)$ be the compression scheme and $\lrabs{\kappa(S)}$ the compression size. Let $\widehat{\err}_{\ell_p}(h;S)$ be the empirical loss of $h$ on $S$ with the $\ell_p$ robust loss.
We can derive the error as follows, 
\begin{align*}
{\err_{\ell_p}}\lr{\rho({\kappa({S})});\cD}
&\overset{(i)}{\lesssim}
\widehat{\err}_{\ell_p}\lr{\rho({\kappa({S})});S}
+
\sqrt{\frac{|\kappa(S)|\log(m)+\log \frac{1}{\delta}}{m}}
\\
&\overset{(ii)}{\lesssim}
\widehat{\err}_{\ell_p}\lr{h^\star;S}
+5\epsilon
+
\sqrt{\frac{|\kappa(S)|\log(m)+\log \frac{1}{\delta}}{m}}
\\
&\overset{(iii)}{\lesssim}
\err_{\ell_p}\lr{h^\star;\cD}
+5\epsilon
+
\sqrt{\frac{|\kappa(S)|\log(m)+\log \frac{1}{\delta}}{m}}
+
\sqrt{\frac{\log \frac{1}{\delta}}{m}}
\\
&\lesssim
\err_{\ell_p}\lr{h^\star;\cD}
+5\epsilon
+
\sqrt{\frac{|\kappa(S)|\log(m)+\log \frac{1}{\delta}}{m}},
\end{align*}
(i) follows from a generalization of sample compression scheme in the agnostic case, see \cref{lem:sample-compression-agnostic},
(ii) follows \cref{eq:adv-compression-lp},
(iii) follows from Hoeffding's inequality.
We take $m$ sufficiently large such that
\begin{align*}
\sqrt{\frac{|\kappa(S)|\log(m)+\log \frac{1}{\delta}}{m}}
\lesssim
\epsilon.
\end{align*}
Re-scale $\epsilon = \epsilon/5$ and plug in the compression size, we get sample complexity of size
\begin{align*}
\cM
=
\cO\lr{
    \frac{1}{\epsilon^2}\lr{
\lrabs{\kappa(S)}\log\frac{1}{\epsilon}+\log\frac{1}{\delta}
}
},
\end{align*}
where $\lrabs{\kappa(S)}$ is upper bounded as follows
\begin{align*}
\cO
\lr{
\frac{1}{\epsilon^3}\fat^3\lr{\cH,c\epsilon/p}\fat^*\lr{\cH,c\epsilon/p}
\log^6\lr{\frac{p^2\,\fat\lr{\cH,\epsilon/8}}{\epsilon^2}}
\log^3\lr{\frac{p}{\epsilon}}\log^2\lr{\frac{m}{\frac{1}{\epsilon}\fat\lr{\cH,c\epsilon/p}\log^2\lr{\frac{p^2\,\fat\lr{\cH,c\epsilon/p}}{\epsilon^2}}}}}.
\end{align*}
We conclude the sample complexity
\begin{align*}
\cM
=
\Tilde{\cO}\lr{
    \frac{1}{\epsilon^5}
\fat^3\lr{\cH,c\epsilon/p}
\fat^*\lr{\cH,c\epsilon/p}
+\frac{1}{\epsilon^2}\log\frac{1}{\delta}
},
\end{align*}
for some numerical constant $c\in (0,\infty)$.
\end{proof}

\section{Proofs for \cref{sec:l1-regression}: Improved Sample Complexity via Median Boosting and Sparsification}\label{app:l1-regression}

\begin{proof}[of \cref{thm:l1-regression}]

Fix $\epsilon,\delta \in (0,1)$ and $p\in [1,\infty]$.
Let $\cH \subseteq [0,1]^\cX$. Fix a distribution $\cD$ over $\cX\times \cY$, and let $S=\lrset{\lr{x_i,y_i}}^m_{i=1}$ be an i.i.d. sample from $\cD$. We first prove for $p\in \lrset{1,\infty}$, and generalize for $p\in (1,\infty)$ by using the Lipschitzness of the $\ell_p$ loss. 
We follow the steps as described in \cref{alg:l1-regression}.

\begin{enumerate}[leftmargin=0.5cm]
    \item Compute ${h^\star}\leftarrow \ell_p\text{-}\RERM_\cH(S)$ in order to get the set of cutoffs $\psi(x,y)=\sup_{z\in\cU(x)}|h^\star(z)-y|$ for $(x,y)\in S$. Let $\psi|_S=\lr{\psi(x_1,y_1),\ldots,\psi(x_m,y_m)}$.
    Our goal is to construct a predictor with an empirical robust loss of $\psi(x,y)^p+\epsilon$, for $p\in (1,\infty)$, and $\psi(x,y)+\epsilon$ for $p\in \lrset{1,\infty}$, for any $(x,y)\in S$, which means that our predictor is an approximate robust ERM.
    \item Define the inflated training data set 
    $$S_\cU = \bigcup_{i
    \in [n]}\lrset{(z,y_{I(z)}):z\in \cU(x_i)},$$ 
    where $I(z) =\min\lrset{i\in [m]:z\in \cU(x_i)}$. For $(z,y)\in S_\cU$, let $\psi(z,y)$ be the $\psi(x,y)$ for which $z \in \cU(x)$ and $y_{I(z)}=y$.
    \item Discretize $S_{\cU}$ to a finite set $\bar{S}_\cU$ as follows. 
    \begin{enumerate}[leftmargin=0.6cm]
        \item Define a set of functions, such that each function
        is defined by an $\epsilon$-approximate $\psi\text{-}\RERM_{\cH}$ optimizer on $d=\cO\lr{\fat\lr{\cH,\epsilon/8}\log^2\lr{\frac{\fat\lr{\cH,\epsilon/8}}{\epsilon^2}}}$ points from $S$,
    $$\hat{\cH}=\lrset{\psi\text{-}\RERM_{\cH}(S',\psi|_{S'},\epsilon): S'\subseteq S, |S'|=d}.$$
    Recall the definition of $\psi\text{-}\RERM_\cH$, see \cref{def:eta-rerm}. 
    
    In order to understand what this definition of $\hat{\cH}$ serves for, see step 4 below.
    The cardinality of this class is bounded as follows
    \begin{align}\label{eq:H-hat}
    |\hat{\cH}|
    \approx
    {m \choose d}
    \lesssim
    \left(\frac{m}{d}\right)^{d}.    
    \end{align}

\item A discretization $\bar{S}_\cU\subseteq S_\cU$ will be defined by covering of the dual class in $\lrnorm{\cdot}_\infty$. Define $\Tilde{\cH}=\hat{\cH}\cup \lrset{h^\star}$.
    Let $L^1_{\Tilde{\cH}}$ be the $L_1$ loss class of $\Tilde{\cH}$, namely, $L^1_{\Tilde{\cH}}=\lrset{\cZ\times\cY \ni (z,y)\mapsto |h(z)-y|: h \in \Tilde{\cH}}$.
    The \textit{dual class} of $L^1_{\Tilde{\cH}}$ ,
    ${L^1_{\Tilde{\cH}}}^* \subseteq \lrbra{0,1}^{\Tilde{\cH}}$,
    is defined as the set of all functions $f_{\lr{z,y}}: \Tilde{\cH} \rightarrow \lrbra{0,1}$ such that $f_{\lr{z,y}}(h) = \Lrabs{h(z)- y}$,
    for any $(z,y)\in S_{\cU}$. 
    Formally, ${L^1_{\Tilde{\cH}}}^*=\lrset{f_{\lr{z,y}}:(z,y)\in S_\cU}$, where $f_{\lr{z,y}}= \lr{f_{(z,y)}(h_1),\ldots,f_{(z,y)}(h_{\lrabs{\Tilde{\cH}}})}$.
    We take $\bar{S}_\cU \subseteq S_\cU$ to be a minimal $\epsilon$-cover for $S_\cU$ in $\lrnorm{\cdot}_\infty$,
    \begin{align}\label{eq:covering-lp}
    \sup_{\lr{z,y}\in S_\cU}\inf_{\lr{\bar{z},\bar{y}}\in \bar{S}_\cU} \lrnorm{f_{\lr{z,y}}-f_{\lr{\bar{z},\bar{y}}}}_\infty\leq \epsilon.  
    \end{align}
    Let $\fat^*\lr{L^1_{\Tilde{\cH}},\epsilon}$ be the dual $\epsilon$-fat-shattering of $L^1_{\Tilde{\cH}}$.
    Applying a covering number argument from \cref{lem:covering-num} on the dual space and upper bounding the dual fat-shattering of the $L_1$ loss class with the dual fat-shattering of $\Tilde{H}$,
    we have the following bound
    \begin{align}\label{eq:covering-l1}
    \begin{split}
    \lrabs{\bar{S}_\cU}
    &=
    \cN\lr{\epsilon,S_\cU,\lrnorm{\cdot}_\infty}
    \\
    &\lesssim 
    \mathrm{exp}\lr{{ \fat^*\lr{L^1_{\Tilde{\cH}},c\epsilon}\log^2\lr{\frac{|\Tilde{\cH}|}{\epsilon}}}}
    \\
    &\lesssim 
    \mathrm{exp}\lr{{ \fat^*\lr{\Tilde{\cH},c\epsilon}\log^2\lr{\frac{|\Tilde{\cH}|}{\epsilon}}}}
    \\
    &\lesssim 
    \mathrm{exp}\lr{{ \fat^*\lr{\cH,c\epsilon}\log^2\lr{\frac{|\Tilde{\cH}|}{\epsilon}}}},
    \end{split}
    \end{align}
    \end{enumerate}
 where $c\in(0,\infty)$ is a numerical constant, derived from the covering argument in \cref{lem:covering-num}.  
    
\item 
Compute a robust variant of the real-valued boosting algorithm $\medboost$ \citep{kegl2003robust,hanneke2019sample} on the discretized set $\bar{S}_\cU$. The output of the algorithm is a uniformly $2\epsilon$-approximate sample compression scheme for the set $\bar{S}_\cU$, for $\approx \log\lr{\lrabs{\bar{S}_\cU}}$ boosting rounds. Moreover, the weak learners are chosen from the set $\hat{\cH}$. Once we have these weak learners, the guarantee of the algorithm follows from \citet[Corollary 6]{hanneke2019sample}. We should explain why we have a weak learner for any distribution over $\bar{S}_\cU$.

\textbf{The existence of weak learners in $\hat{\cH}$.}
Let $d=\cO\lr{\fat\lr{\cH,\epsilon/8}\log^2\lr{\frac{\fat\lr{\cH,\epsilon/8}}{\epsilon^2}}}$ and let $\psi(\bar{z},\bar{y})$ be the $\psi(x,y)$ for which $\bar{z} \in \cU(x)$. 
Taking $\delta=1/3$, we know that for any distribution $\cP$ on $\bar{S}_\cU$, upon receiving an i.i.d. sample $S''$ from $\cP$ of size $d$, with probability $2/3$ over sampling $S''$ from $\cP$, 
for any ${h}\in\cH$ satisfying $\forall (\bar{z},\bar{y})\in S'':\; \lrabs{{h}(
\bar{z})-\bar{y}}\leq \psi(\bar{z},\bar{y})+\epsilon$, 
it holds that $\hP_{(\bar{z},\bar{y})\sim\cP}\lr{(\bar{z},\bar{y}): \lrabs{h(\bar{z})-\bar{y}}>\psi(\bar{z},\bar{y})+2\epsilon}\leq 1/3$.
That is, such a function is a $\lr{2\epsilon,1/6}$-weak learner for $\cP$ and $h^\star$.
We can conclude that for any distribution $\cP$ on $\bar{S}_\cU$, there exists a set of points $S''\subseteq \bar{S}_\cU$ of size $d$ that defines a weak learner for $\cP$ and $h^\star$.

Furthermore, we can find these weak learners in $\hat{\cH}$ as follows.
Let $S'$ be the $d$ points in $S$ that the perturbed points $S''$ originated from.
That is, $S''\subseteq \bigcup_{(x,y)\in {S'}}\bigcup \lrset{(\bar{z},\bar{y}): \bar{z}\in \cU(x)}$.
Therefore, we can conclude that $\hat{h} = \psi\text{-}\RERM_\cH(S',\psi|_{S'},\epsilon)$ is a weak learner, and can be found in $\hat{\cH}$.
So, we can think of $\hat{\cH}$ as a pool of weak learners for any possible distribution over the discretized set $\bar{S_\cU}$.

    \begin{algorithm}[H]
    \caption{Robust $\medboost$}\label{alg:med-boost}
    \textbf{Input:} $\cH, S, \bar{S}_\cU.$\\
    \textbf{Parameters:}  Approximation parameter $\epsilon\in (0,1)$, number of boosting rounds $T\geq1$, weak learner sample size $d\geq 1$, loss parameter $p\in [1,\infty]$, cutoff parameters $\psi|_S=\lr{\psi(x_1,y_1),\ldots,\psi(x_m,y_m)}$ for $(x_i,y_i),
\in S$ and $\psi(\bar{z},\bar{y})$ is the $\psi(x,y)$ for which $\bar{z} \in \cU(x)$.\\
    \textbf{Algorithms used:} $\epsilon$-approximate $\psi$-robust empirical risk minimizer $\psi\text{-}\RERM_\cH$ (\cref{def:eta-rerm}).\\
    \textbf{Initialize} $\cP_1$ = Uniform($\bar{S}_\cU$).\\
     For $t=1,\ldots,T$:
        \begin{enumerate}

            \item[]\textcolor{DarkBlue}{$\triangleright$ Compute a weak base learner w.r.t. distribution $\cP_t$ by finding $d$ points in $S$ and executing $\psi\text{-}\RERM_\cH$ on them.}
            \item 
            Find $d$ points ${S''_t}\subseteq \bar{S}_\cU$ such that any $h\in \cH$ satisfying:
            $\forall \lr{\bar{z},\bar{y}}\in S''_t:\; \lrabs{{h}(\bar{z})-\bar{y}}^p\leq \psi(\bar{z},\bar{y})^p+\epsilon$, 
            it holds that 
            $\hE_{(\bar{z},\bar{y})\sim\cP_t} \LRbra{\hI \LRset{\lrabs{h(\bar{z})- \bar{y}}^p \geq \psi(\bar{z},\bar{y})^p+2\epsilon}}\leq 1/3$.
            (See the analysis for why this set exists).

            \item Let $S'_t$ be the $d$ points in $S$ that 
            $S''_t$ originated from. Formally, $S''_t\subseteq \bigcup_{(x,y)\in {S'_t}}\bigcup \lrset{(\bar{z},\bar{y}): \bar{z}\in \cU(x)}$.
            \item Compute $\hat{h}_t = \psi\text{-}\RERM_\cH(S'_t,\psi|_{S'_t},\epsilon)$. From steps (a) and (b), it follows that $\hat{h}_t$ is a $\lr{2\epsilon,1/6}$-weak learner with respect to the distribution $\cP_t$ over $\bar{S}_\cU$.
            \item[]\textcolor{DarkBlue}{$\triangleright$ Update the weight of the weak learner in the ensemble and make a multiplicative weights update on $\cP_t$.}
            \item For $i=1,\ldots,n=\lrabs{\bar{S}_\cU}$:\\
                    Set $$w^{(t)}_i = 1-2\hI\Lrbra{\lrabs{\hat{h}_t(\bar{z_i})-\bar{y_i}}^p>\psi(\bar{z_i},\bar{y_i})^p+2\epsilon}.$$
            \item Set $$\alpha_t=\frac{1}{2}\log\lr{\frac{\lr{1-1/6}\sum^n_{i=1}\cP_t\lr{\bar{z_i},\bar{y_i}}\hI\lrbra{w_i^{(t)}=1}}{\lr{1+1/6}\sum^n_{i=1}\cP_t\lr{\bar{z_i},\bar{y_i}}\hI\lrbra{w_i^{(t)}=-1}}}.$$
            \item If $\alpha_t=\infty$: return $T$ copies of $h_t$, $\lr{\alpha_1=1,\ldots,\alpha_T=1}$, and $S'_t$. \\
            Else: $$P_{t+1}(\bar{z_i},\bar{y_i}) = P_{t}(\bar{z_i},\bar{y_i}) \frac{\exp\lr{-\alpha_t w_i^{t}}}{\sum^n_{j=1}\cP_t\lr{\bar{z_j},\bar{y_j}}\exp\lr{-\alpha_t w_j^{t}}}.$$

        \end{enumerate}
    \textbf{Output:} Hypotheses $\hat{h}_1,\ldots,\hat{h}_T$, coefficients $\alpha_1,\ldots,\alpha_T$ and sets $S'_1,\ldots,S'_T$.
\end{algorithm}

\noindent\textbf{A uniformly $4\epsilon$-approximate adversarially robust sample compression scheme for $S$.} 
The output of $\medboost$ is a uniformly $2\epsilon$-approximate sample compression scheme for the set $\bar{S}_\cU$. We show that this is a uniformly $4\epsilon$-approximate \emph{adversarially robust} sample compression scheme for $S$, that is, a sample compression for $S$ scheme with respect to the robust loss.

For $T \approx \log\lrabs{\bar{S}_\cU}$ boosting rounds, it follows from \citet[Corollary 6]{hanneke2019sample} that the output of the algorithm satisfy
\begin{align}\label{eq:Su-bar-median-l1-reg}
\forall \lr{\bar{z},\bar{y}}\in \bar{S}_\cU:\;\lrabs{\med\Lr{\hat{h}_1(\bar{z}),\ldots,\hat{h}_T(\bar{z});\alpha_1,\ldots,\alpha_T}-\bar{y}}\leq \psi\lr{\bar{z},\bar{y}}+2\epsilon,
\end{align}
$\med\Lr{\hat{h}_1(\bar{z}),\ldots,\hat{h}_T(\bar{z});\alpha_1,\ldots,\alpha_T}$ is the weighted median of $\hat{h}_1,\ldots,\hat{h}_T$ with weights $\alpha_1,\ldots,\alpha_T$.
From the covering argument (\cref{eq:covering-l1}), this implies that
\begin{align}\label{eq:Su-median-l1-reg}
\forall \lr{z,y}\in S_\cU:\;\lrabs{\med\Lr{\hat{h}_1(z),\ldots,\hat{h}_T(z);\alpha_1,\ldots,\alpha_T}-y}\leq \psi\lr{z,y} + 4\epsilon.
\end{align}
%
Indeed, for any $\lr{z,y}\in S_\cU$ there exists $\lr{\bar{z},\bar{y}}\in \bar{S}_\cU$, such that for any 
$h\in \Tilde{\cH}$,
\begin{align*}
\LRabs{\Lrabs{h(z)-y}-\Lrabs{h(\bar{z})-\bar{y}}}\leq \epsilon.
\end{align*}
Specifically, it holds for 
$\lrset{\hat{h}_1,\ldots,\hat{h}_T}\subseteq \Tilde{\cH}$ and $h^\star\in\Tilde{\cH}$.
So, 
\begin{align}\label{eq:Su-median-l1-reg}
\begin{split}
\LRabs{\med\Lr{\hat{h}_1(z),\ldots,\hat{h}_T(z);\alpha_1,\ldots,\alpha_T}-y}
    &\overset{(a)}{=}
    \LRabs{\med\Lr{\hat{h}_1(z)-y,\ldots,\hat{h}_T(z)-y;\alpha_1,\ldots,\alpha_T}}
    \\
    &\overset{(b)}{\leq}
    \LRabs{\med\Lr{\hat{h}_1(\bar{z})-\bar{y},\ldots,\hat{h}_T(\bar{z})-\bar{y};\alpha_1,\ldots,\alpha_T}} + \epsilon
    \\
    &\overset{(c)}{=}
    \LRabs{\med\Lr{\hat{h}_1(\bar{z}),\ldots,\hat{h}_T(\bar{z});\alpha_1,\ldots,\alpha_T}-\bar{y}} + \epsilon
    \\
    &\overset{(d)}{\leq}
    \lrabs{h^\star(\bar{z})-\bar{y}}+3\epsilon
    \\
    &\overset{(e)}{\leq}
    \lrabs{h^\star(z)-y}+4\epsilon
    \\
    &\overset{(f)}{=}
    \psi(z,y)+4\epsilon,
\end{split}
\end{align}
$(a)\text{+}(c)$ follow since the median is translation invariant,
$(b)\text{+}(e)$ follow from the covering argument,
$(d)$ holds since the returned function by $\medboost$ is a uniformly $2\epsilon$-approximate sample compression for $\bar{S}_\cU$,
$(f)$ follows from the definition in step 2 of the algorithm.

Finally, from \cref{eq:Su-median-l1-reg} we conclude a uniformly $4\epsilon$-approximate \emph{adversarially robust} sample compression scheme for $S$,
\begin{align}\label{eq:S-median-l1}
\forall \lr{x,y}\in S:\;\sup_{z\in\cU(x)}\lrabs{\med\Lr{\hat{h}_1(z),\ldots,\hat{h}_T(z);\alpha_1,\ldots,\alpha_T}-y}\leq \psi(x,y)+4\epsilon.
\end{align}

\paragraph{Bounding the compression size.} 
We have $T=\cO\lr{\log\lrabs{\bar{S}_\cU}}$ hypotheses, where each one is representable by $d=\cO\lr{\fat\lr{\cH,\epsilon/8}\log^2\lr{\frac{\fat\lr{\cH,\epsilon/8}}{\epsilon^2}}}$ points. By counting the number of predictors using \cref{eq:covering-l1}, we get
\begin{align*}
    \begin{split}
    \log\lr{\lrabs{\bar{S}_\cU}}
    &\lesssim
    \fat^*\lr{\cH,c\epsilon}\log^2\lr{\frac{|\Tilde{\cH}|}{\epsilon}}
    \\
    &\lesssim 
    \fat^*\lr{\cH,c\epsilon}\log^2\lr{\frac{1}{\epsilon}\lr{\lr{\frac{m}{d}}^d+1}}.
    \end{split}
    \end{align*}
We have a compression of size $\cO\lr{d\log\lr{ \lrabs{\bar{S}_\cU}}}$, which is already sufficient for deriving generalization. 
We can reduce further the number of predictors to be  \emph{independent} of the sample size, thereby reducing the sample compression size and improving the sample complexity.
\item We follow the sparsification method suggested by \citet{hanneke2019sample}. 
The idea is that by sampling functions from the ensemble, we can guarantee via a uniform convergence in the dual space, that it is sufficient to have roughly $\approx \fat^*(\cH,c\epsilon)$ predictors.

For $\alpha_1,\ldots,\alpha_T\in [0,1]$ with $\sum^T_{t=1}\alpha_t = 1$, we denote the categorical distribution by $\mathrm{Cat}\lr{\alpha_1,\ldots,\alpha_T}$, which is a discrete distribution on the set $[T]$ with probability of $\alpha_t$ on $t\in [T]$.
The inputs to the algorithm are $\tau(x,y)=\psi(x,y)^p+5\epsilon$ and $k=\cO\lr{\fat^*\lr{\cH,c\epsilon}\log^2\lr{\fat^*\lr{\cH,c\epsilon}/\epsilon}}$, where $c\in(0,\infty)$ is a numerical constant.

    \begin{algorithm}[H]
    \caption{Sparsify}\label{alg:Sparsify}
    \textbf{Input:} Hypotheses $\hat{h}_1,\ldots,\hat{h}_T$, coefficients $\alpha_1,\ldots,\alpha_T$, $S=\lrset{(x_i,y_i}^m_{i=1}$.\\
    \textbf{Parameter:} Number of functions to sample $k\geq 1$, cutoff parameters $\lr{\tau(x_1,y_1),\ldots,\tau(x_m,y_m)}$.
    \begin{enumerate}
        \item Let $\alpha'_t=\alpha_t\big/ \sum^T_{s=1}\alpha_s$.
        \item \textbf{Repeat}:
        \begin{enumerate}
            \item Sample $\lr{J_1,\ldots,J_k}\sim \mathrm{Cat\lr{\alpha'_1,\ldots,\alpha'_T}^k}$.
            \item Let $\cF=\lrset{h_{J_1},\ldots,h_{J_k}}$.
            \item \textbf{Until} $\forall (x,y)\in S:\;
          \lrabs{\lrset{f\in\cF:\sup_{{z\in\cU(x)}}\lrabs{f(z)-y}^p>\tau(x,y)}}<k/2$.
        \end{enumerate}
    \end{enumerate}
    \textbf{Output:} Hypotheses $\hat{h}_{J_1},\ldots,\hat{h}_{J_k}$.
\end{algorithm}

The sparsification method returns set of functions $\lrset{\hat{h}_{J_1},\ldots,\hat{h}_{J_k}}$, such that 
\begin{align}
\forall \lr{x,y}\in S:\;\sup_{z\in\cU(x)}\Lrabs{\med\Lr{\hat{h}_{J_1}(x),\ldots,\hat{h}_{J_k}(x)}-y}\leq \psi(x,y)+5\epsilon.
\end{align}
We get a uniformly $5\epsilon$-approximate \emph{adversarially robust} sample compression scheme for $S$, where we have $k=\cO\lr{\fat^*\lr{\cH,c\epsilon}\log^2\lr{\fat^*\lr{\cH,c\epsilon}/\epsilon}}$ functions, and each function is representable by 
$d=\cO\lr{\fat\lr{\cH,\epsilon/8}\log^2\lr{\fat\lr{\cH,\epsilon/8}/\epsilon^2}}$ points.

\paragraph{Encoding weak learners.} Each weak learner is encoded by a multiset $S'\subseteq S$ of size $d$ and is constructed by computing some $\hat{h}\in \cH$ that solves the  constrained optimization 
$$ \sup_{z\in\cU(x)}\lrabs{\hat{h}(z)-y}\leq \psi(x,y)+\epsilon, \; \forall (x,y)\in S'.$$
We encode each $\psi(x,y)$ by some approximation $\Tilde{\psi}(x,y)$, such that $\lrabs{\Tilde{\psi}(x,y)-\psi(x,y)}\leq \epsilon$, by discretizing $[0,1]$ to $1/\epsilon$ buckets of size $\epsilon$, and each $\psi(x,y)$ is rounded down to the closest value $\Tilde{\psi}(x,y)$. Each approximation requires to encode $\log\lr{1/\epsilon}$ bits, and so each learner encodes $d\log\lr{1/\epsilon}$ bits and $d$ samples. We have $k$ weak learners, and the compression size is 
\begin{align*}
    k(d+d\log\lr{1/\epsilon})
    &\leq
    2kd\log\lr{1/\epsilon}.
\end{align*}
Therefore, we have a uniform $6\epsilon$-approximate compression for $\ell_1$ and $\ell_\infty$ losses of size
\begin{align*}
    \cO\lr{\fat\lr{\cH,\epsilon/8}\fat^*\lr{\cH,c\epsilon}\log^2\lr{\frac{\fat\lr{\cH,\epsilon/8}}{\epsilon^2}}
    \log^2\lr{\frac{\fat^*\lr{\cH,c\epsilon}}{\epsilon}}\log\frac{1}{\epsilon}}.
\end{align*}
\end{enumerate}
\paragraph{Generalizing for $p\in(1,\infty)$.} 
We rely on the Lipschitzness of the $\ell_p$ loss and rescaling the approximation parameter $\epsilon$ to $\epsilon/p$.

Recall the covering of $S_\cU$ in step 3(b).
Note that an $(\epsilon/p)$-cover for the $L_1$ loss class is an $\epsilon$-cover for the $L_p$ loss class due to the Lipschitzness of the $\ell_p$ loss
    \begin{align*}
        \LRabs{\lrabs{h(z)-y}^p-\lrabs{h(\bar{z})-\bar{y}}^p}
        &\leq
        p\LRabs{\lrabs{h(z)-y}-\lrabs{h(\bar{z})-\bar{y}}}
        \\
        &\leq
        p\epsilon.
    \end{align*}

Moreover, we constructed a function $f=\med\Lr{\hat{h}_{J_1},\ldots,\hat{h}_{J_k}}$ with $\sup_{z\in\cU(x)}\lrabs{f(z)-y} \leq \psi(x,y)+\epsilon$ for any $(x,y)\in S$. Note that since $\lrabs{f(\cdot)-y}\in [0,1]$, the same $z$ that maximizes the $\ell_1$ loss also maximizes for any $\ell_p$.
This implies that 
\begin{align*}
\sup_{z\in\cU(x)}\lrabs{f(z)-y}^p 
\overset{(i)}{\leq} 
\lr{\lr{\psi(x,y)}+\epsilon}^p 
\overset{(ii)}{\leq}
\psi(x,y)^p+p\epsilon,
\end{align*}
where (i) follows by just raising both sides to the power of $p$ and
(ii) follows since the function $x\mapsto \lrabs{x-y}^p$ is $p$-Lipschitz for $(x-y)\in [0,1]$, and so
\begin{align*}
    \lrabs{\lr{\psi(x,y)+\epsilon}^p - \psi(x,y)^p}
    &\leq
    p\lrabs{\psi(x,y)+\epsilon - \psi(x,y)}
    \\
    &\leq
    p\epsilon.
\end{align*}
By rescaling $p\epsilon$ to $\epsilon$, we get 
\begin{align*}
        \sup_{z\in\cU(x)}\lrabs{\med\lr{\hat{h}_{J_1}(z),\ldots,\hat{h}_{J_k}(z)}-y}^p 
        &\leq 
        \psi(x,y)^p+\epsilon,
\end{align*}
%
%
where the number of functions in the ensemble is 
$$k=\cO\lr{\fat^*\lr{\cH,c\epsilon/p}\log^2\lr{\frac{p\,\fat^*\lr{\cH,c\epsilon/p}}{\epsilon}}},$$
and each function is represented by a set of samples of size 
$$d= \cO\lr{\fat\lr{\cH,c\epsilon/p}\log^2\lr{\frac{p^2\,\fat\lr{\cH,c\epsilon/p}}{\epsilon^2}}}.$$ 
Therefore, we have an approximate compression for $\ell_p$ of size
\begin{align*}
\cO\lr{\fat\lr{\cH,c\epsilon/p}\fat^*\lr{\cH,c\epsilon/p}\log^2\lr{\frac{p^2\,\fat\lr{\cH,c\epsilon/p}}{\epsilon^2}}
    \log^2\lr{\frac{p\,\fat^*\lr{\cH,c\epsilon/p}}{\epsilon}}\log\frac{p}{\epsilon}}.
\end{align*}
\paragraph{Generalization bound.}
Let $(\kappa,\rho)$ be the compression scheme and $\lrabs{\kappa(S)}$ the compression size. Let $\widehat{\err}_{\ell_p}(h;S)$ be the empirical loss of $h$ on $S$ with the $\ell_p$ robust loss.
We can derive the error as follows, 
\begin{align*}
{\err_{\ell_p}}\lr{\rho({\kappa({S})});\cD}
&\overset{(i)}{\lesssim}
\widehat{\err}_{\ell_p}\lr{\rho({\kappa({S})});S}
+
\sqrt{\frac{|\kappa(S)|\log(m)+\log \frac{1}{\delta}}{m}}
\\
&\overset{(ii)}{\lesssim}
\widehat{\err}_{\ell_p}\lr{h^\star;S}
+6\epsilon
+
\sqrt{\frac{|\kappa(S)|\log(m)+\log \frac{1}{\delta}}{m}}
\\
&\overset{(iii)}{\lesssim}
\err_{\ell_p}\lr{h^\star;\cD}
+6\epsilon
+
\sqrt{\frac{|\kappa(S)|\log(m)+\log \frac{1}{\delta}}{m}}
+
\sqrt{\frac{\log \frac{1}{\delta}}{m}}
\\
&\lesssim
\err_{\ell_p}\lr{h^\star;\cD}
+6\epsilon
+
\sqrt{\frac{|\kappa(S)|\log(m)+\log \frac{1}{\delta}}{m}},
\end{align*}
(i) follows from a generalization of sample compression scheme in the agnostic case, see \cref{lem:sample-compression-agnostic},
(ii) follows from the approximate sample compression we proved above,
(iii) follows from Hoeffding's inequality.
We take $m$ sufficiently large such that
\begin{align*}
\sqrt{\frac{|\kappa(S)|\log(m)+\log \frac{1}{\delta}}{m}}
\lesssim
\epsilon.
\end{align*}
Re-scale $\epsilon = \epsilon/6$ and plug in the compression size, we get sample complexity of size
\begin{align*}
\cM
=
\cO\lr{
    \frac{1}{\epsilon^2}\lr{
\fat\lr{\cH,c\epsilon/p}\fat^*\lr{\cH,c\epsilon/p}\log^2\lr{\frac{p^2\,\fat\lr{\cH,c\epsilon/p}}{\epsilon^2}}
    \log^2\lr{\frac{p\,\fat^*\lr{\cH,c\epsilon/p}}{\epsilon}}\log^2\lr{\frac{p}{\epsilon}}+\log\frac{1}{\delta}
}
},
\end{align*}
for some numerical constant $c\in (0,\infty)$.
\end{proof}

\section{Proofs for \cref{sec:eta-beta-regression}: Robust $(\eta,\beta)$-Regression}\label{app:eta-beta-regression}

\subsection{Realizable}

\begin{proof}[of \cref{thm:re-uni-reg}]\label{prf:re-eta-beta-reg}
Fix $\epsilon,\delta,\beta,\eta \in (0,1)$.
Let $\cH \subseteq [0,1]^\cX$. Fix a distribution $\cD$ over $\cX\times \cY$ that is $\eta$-uniformly realizable by $\cH$, and let $S=\lrset{\lr{x_i,y_i}}^m_{i=1}$ be an i.i.d. sample from $\cD$.

We elaborate on each one of the steps as described in \cref{alg:improper-robust-learner-eta-beta}.

\begin{enumerate}[leftmargin=0.5cm]
    \item Define the inflated training data set 
    $$S_\cU = \bigcup_{i
    \in [n]}\lrset{(z,y_{I(z)}):z\in \cU(x_i)},$$ 
    where $I(z) =\min\lrset{i\in [m]:z\in \cU(x_i)}$. 
    
    \item Discretize $S_{\cU}$ to a finite set $\bar{S}_\cU$ as follows. 
    \begin{enumerate}[leftmargin=0.6cm]
        \item Define a set of functions, such that each function is defined by an $\beta$-approximate $\psi$--$\RERM_\cH$ optimizer on $d=\cO\lr{\fat\lr{\cH,\beta/8}\log^2\lr{\frac{\fat\lr{\cH,\beta/8}}{\beta}}}$ points from $S$,  with fixed cutoff parameters $\psi(x,y)=\eta$ for each $(x,y)\in S$, 
    $$\hat{\cH}=\lrset{\psi\text{-}\RERM_{\cH}(S',\psi|_{S'},\beta): S'\subseteq S, |S'|=d}.$$
    Recall the definition of $\psi\text{-}\RERM_\cH$, see \cref{def:eta-rerm}.
    The cardinality of this class is bounded as follows
    \begin{align}\label{eq:H-hat}
    |\hat{\cH}|
    \approx
    {m \choose d}
    \lesssim
    \left(\frac{m}{d}\right)^{d}.   
    \end{align}

 \item A discretization $\bar{S}_\cU\subseteq S_\cU$ will be defined by covering of the dual class in $\lrnorm{\cdot}_\infty$. Define $\Tilde{\cH}=\hat{\cH}\cup \lrset{h^\star}$.
    Let $L^1_{\Tilde{\cH}}$ be the $L_1$ loss class of $\Tilde{\cH}$, namely, $L^1_{\Tilde{\cH}}=\lrset{\cZ\times\cY \ni (z,y)\mapsto |h(z)-y|: h \in \Tilde{\cH}}$.
    The \textit{dual class} of $L^1_{\Tilde{\cH}}$ ,
    ${L^1_{\Tilde{\cH}}}^* \subseteq \lrbra{0,1}^{\Tilde{\cH}}$,
    is defined as the set of all functions $f_{\lr{z,y}}: \Tilde{\cH} \rightarrow \lrbra{0,1}$ such that $f_{\lr{z,y}}(h) = \Lrabs{h(z)- y}$,
    for any $(z,y)\in S_{\cU}$. 
    Formally, ${L^1_{\Tilde{\cH}}}^*=\lrset{f_{\lr{z,y}}:(z,y)\in S_\cU}$, where $f_{\lr{z,y}}= \lr{f_{(z,y)}(h_1),\ldots,f_{(z,y)}(h_{\lrabs{\Tilde{\cH}}})}$.
    We take $\bar{S}_\cU \subseteq S_\cU$ to be a minimal $\beta$-cover for $S_\cU$ in $\lrnorm{\cdot}_\infty$,
    \begin{align}\label{eq:covering-lp}
    \sup_{\lr{z,y}\in S_\cU}\inf_{\lr{\bar{z},\bar{y}}\in \bar{S}_\cU} \lrnorm{f_{\lr{z,y}}-f_{\lr{\bar{z},\bar{y}}}}_\infty\leq \beta.  
    \end{align}
    Let $\fat^*\lr{L^1_{\Tilde{\cH}},\beta}$ be the dual $\beta$-fat-shattering of $L^1_{\Tilde{\cH}}$.
    Applying a covering number argument from \cref{lem:covering-num} on the dual space and upper bounding the dual fat-shattering of the $L_1$ loss class with the dual fat-shattering of $\Tilde{H}$,
    we have the following bound
    \begin{align}\label{eq:Su-covering-cutoff}
    \begin{split}
    \lrabs{\bar{S}_\cU}
    &=
    \cN\lr{\beta,S_\cU,\lrnorm{\cdot}_\infty}
    \\
    &\lesssim 
    \mathrm{exp}\lr{{ \fat^*\lr{L^1_{\Tilde{\cH}},c\beta}\log^2\lr{\frac{|\Tilde{\cH}|}{\beta}}}}
    \\
    &\lesssim 
    \mathrm{exp}\lr{{ \fat^*\lr{\Tilde{\cH},c\beta}\log^2\lr{\frac{|\Tilde{\cH}|}{\beta}}}}
    \\
    &\lesssim 
    \mathrm{exp}\lr{{ \fat^*\lr{\cH,c\beta}\log^2\lr{\frac{|\Tilde{\cH}|}{\beta}}}},
    \end{split}
    \end{align}
    \end{enumerate}
 where $c\in(0,\infty)$ is a numerical constant, derived from the covering argument in \cref{lem:covering-num}. 
    
\item Compute a robust version of the real-valued boosting algorithm $\medboost$ (\cref{alg:med-boost}) on the discretized set $\bar{S}_\cU$. The inputs to the algorithm are as follows. Set $\epsilon=\beta$, $\psi|_S=\lr{\eta,\ldots,\eta}$, $p=1$, and $T \approx \log\lr{\lrabs{\bar{S}_\cU}}$ rounds of boosting.

The output of the algorithm is a uniform $\beta$-approximate sample compression scheme for the set $\bar{S}_\cU$. Moreover, the weak learners are chosen from the set $\hat{\cH}$. Once we have these weak learners, the guarantee of the algorithm follows from \citet[Corollary 6]{hanneke2019sample}. We should explain why we have a weak learner for any distribution over $\bar{S}_\cU$.

\textbf{The existence of weak learners in $\hat{\cH}$.}
From \cref{thm:generalization-interpolation}, taking $\epsilon=\delta=1/3$, we know that for any distribution $\cP$ on $\bar{S}_\cU$, upon receiving an i.i.d. sample $S''$ from $\cP$ of size \\ $\cO\lr{\fat\lr{\cH,\beta/8}\log^2\lr{\frac{\fat\lr{\cH,\beta/8}}{\beta}}}$, with probability $2/3$ over sampling $S''$ from $\cP$, 
for any ${h}\in\cH$ with $\forall (z,y)\in S'':\; \lrabs{{h}(z)-y}\leq \eta$, 
it holds that $\hP_{(z,y)\sim\cP}\lrset{(z,y): \lrabs{h(z)-y}>\eta+\beta}\leq 1/3$. That is, such a function is a $\lr{\beta,1/6}$-weak learner for $\cP$ (see \cref{def:weak-learner}). We can conclude that for any distribution $\cP$ on $\bar{S}_\cU$, there exists a set of points $S''\subseteq \bar{S}_\cU$ of size $\cO\lr {\fat\lr{\cH,\beta/8}\log^2\lr{\frac{\fat\lr{\cH,\beta/8}}{\beta}}}$ that defines a weak learner for $\cP$.

Moreover, we can find these weak learners in $\hat{\cH}$ as follows.
Let $S'$ be the 
$\cO\lr {\fat\lr{\cH,\beta/8}\log^2\lr{\frac{\fat\lr{\cH,\beta/8}}{\beta}}}$ points in $S$ that the perturbed points $S''$ originated from.
That is, $S''\subseteq \bigcup_{(x,y)\in {S'}}\bigcup \lrset{(z,y): z\in \cU(x)}$.
Therefore, we can conclude that $\hat{h} = \psi\text{-}\RERM_\cH(S',\psi|_{S'},\beta)$ is a weak learner, and can be found in $\hat{\cH}$.
So, we can think of $\hat{\cH}$ as a pool of weak learners for any possible distribution over the discretized set $\bar{S_\cU}$. 

\noindent\textbf{A uniformly $3\beta$-approximate adversarially robust sample compression scheme for $S$.} 
The output of $\medboost$ is a uniformly $\beta$-approximate sample compression scheme for the set $\bar{S}_\cU$. We show that this is a uniformly $3\beta$-approximate \emph{adversarially robust} sample compression scheme for $S$, that is, a sample compression for $S$ scheme with respect to the robust loss. 

For $T \approx \log\lrabs{\bar{S}_\cU}$ boosting rounds, it follows from \citet[Corollary 6]{hanneke2019sample} that the output of the algorithm satisfy
\begin{align}\label{eq:Su-bar-median-l1-reg}
\forall \lr{\bar{z},\bar{y}}\in \bar{S}_\cU:\;\lrabs{\med\Lr{\hat{h}_1(\bar{z}),\ldots,\hat{h}_T(\bar{z});\alpha_1,\ldots,\alpha_T}-\bar{y}}\leq \eta+\beta,
\end{align}
$\med\Lr{\hat{h}_1(\bar{z}),\ldots,\hat{h}_T(\bar{z});\alpha_1,\ldots,\alpha_T}$ is the weighted median of $\hat{h}_1,\ldots,\hat{h}_T$ with weights $\alpha_1,\ldots,\alpha_T$.
From the covering argument (\cref{eq:Su-covering-cutoff}), this implies that
\begin{align}\label{eq:Su-median-l1-reg-cutoff}
\forall \lr{z,y}\in S_\cU:\;\lrabs{\med\Lr{\hat{h}_1(z),\ldots,\hat{h}_T(z);\alpha_1,\ldots,\alpha_T}-y}\leq \eta + 3\beta.
\end{align}
Indeed, for any $\lr{z,y}\in S_\cU$ there exists $\lr{\bar{z},\bar{y}}\in \bar{S}_\cU$, such that for any 
$h\in \Tilde{\cH}$,
\begin{align*}
\LRabs{\Lrabs{h(z)-y}-\Lrabs{h(\bar{z})-\bar{y}}}\leq \beta.
\end{align*}
Specifically, it holds for 
$\lrset{\hat{h}_1,\ldots,\hat{h}_T}\subseteq \Tilde{\cH}$ and $h^\star\in\Tilde{\cH}$.
So, 
\begin{align}\label{eq:Su-median-l1-reg-cutoff}
\begin{split}
\LRabs{\med\Lr{\hat{h}_1(z),\ldots,\hat{h}_T(z);\alpha_1,\ldots,\alpha_T}-y}
    &\overset{(a)}{=}
    \LRabs{\med\Lr{\hat{h}_1(z)-y,\ldots,\hat{h}_T(z)-y;\alpha_1,\ldots,\alpha_T}}
    \\
    &\overset{(b)}{\leq}
    \LRabs{\med\Lr{\hat{h}_1(\bar{z})-\bar{y},\ldots,\hat{h}_T(\bar{z})-\bar{y};\alpha_1,\ldots,\alpha_T}} + \beta
    \\
    &\overset{(c)}{=}
    \LRabs{\med\Lr{\hat{h}_1(\bar{z}),\ldots,\hat{h}_T(\bar{z});\alpha_1,\ldots,\alpha_T}-\bar{y}} + \beta
    \\
    &\overset{(d)}{\leq}
    \lrabs{h^\star(\bar{z})-\bar{y}}+2\beta
    \\
    &\overset{(e)}{\leq}
    \lrabs{h^\star(z)-y}+3\beta
    \\
    &\overset{(f)}{=}
    \eta+3\beta,
\end{split}
\end{align}
$(a)\text{+}(c)$ follow since the median is translation invariant,
$(b)\text{+}(e)$ follow from the covering argument,
$(d)$ holds since the returned function by $\medboost$ is a uniformly $\beta$-approximate sample compression for $\bar{S}_\cU$,
$(f)$ follows from the assumption of $\eta$-realizability.

Finally, from \cref{eq:Su-median-l1-reg-cutoff} we conclude a uniformly $3\beta$-approximate \emph{adversarially robust} sample compression scheme for $S$,
\begin{align}\label{eq:S-median-l1}
\forall \lr{x,y}\in S:\;\sup_{z\in\cU(x)}\lrabs{\med\Lr{\hat{h}_1(z),\ldots,\hat{h}_T(z);\alpha_1,\ldots,\alpha_T}-y}\leq \eta+3\beta.
\end{align}

\paragraph{Bounding the compression size.} 
We have $T=\cO\lr{\log\lrabs{\bar{S}_\cU}}$ hypotheses, where each one is representable by $d=\cO\lr{\fat\lr{\cH,\beta/8}\log^2\lr{\frac{\fat\lr{\cH,\beta/8}}{\beta^2}}}$ points. By counting the number of predictors using \cref{eq:Su-covering-cutoff}, we get
\begin{align*}
    \begin{split}
    \log\lr{\lrabs{\bar{S}_\cU}}
    &\lesssim
    \fat^*\lr{\cH,c\beta}\log^2\lr{\frac{|\Tilde{\cH}|}{\beta}}
    \\
    &\lesssim 
    \fat^*\lr{\cH,c\beta}\log^2\lr{\frac{1}{\beta}\lr{\lr{\frac{m}{d}}^d+1}}.
    \end{split}
    \end{align*}
We have a compression of size $\cO\lr{d\log\lr{ \lrabs{\bar{S}_\cU}}}$, which is already sufficient for deriving generalization. 
We can reduce further the number of predictors to be  \emph{independent} of the sample size, thereby reducing the sample compression size and improving the sample complexity.

\item Compute the sparsification method (\cref{alg:Sparsify}).
The idea is that by sampling functions from the ensemble, we can guarantee via a uniform convergence for the dual space, that with high probability it is sufficient to have roughly $\approx \fat^*(\cH,\cO(\beta))$ predictors.
Applying \citet[Theorem 10]{hanneke2019sample} with the parameters $\tau(x,y)=\eta+4\beta$ and $k=\cO\lr{\fat^*\lr{\cH,c\beta}\log^2\lr{\fat^*\lr{\cH,c\beta}/\beta}}$, where $c\in(0,\infty)$ is a numerical constant.
The sparsification method returns set of functions $\lrset{\hat{h}_{J_1},\ldots,\hat{h}_{J_k}}$, such that 
\begin{align*}
\forall \lr{x,y}\in S:\;\sup_{z\in\cU(x)}\Lrabs{\med\Lr{\hat{h}_{J_1}(x),\ldots,\hat{h}_{J_k}(x)}-y}\leq \eta+4\beta.
\end{align*}

We get a uniformly $4\beta$-approximate \emph{adversarially robust} sample compression scheme for $S$, where we have $\cO\lr{\fat^*\lr{\cH,c\beta}\log^2\lr{\fat^*\lr{\cH,c\beta}/\beta^2}}$ functions, and each function is representable by \\ $\cO\lr{\fat\lr{\cH,\beta/8}\log^2\lr{\fat\lr{\cH,\beta/8}/\beta^2}}$ points.
\paragraph{Encoding weak learners.} Each weak learner is encoded by a multiset $S'\subseteq S$ of size $d$ and is constructed by computing some $\hat{h}\in \cH$ that solves the  constrained optimization 
$$ \sup_{z\in\cU(x)}\lrabs{\hat{h}(z)-y}\leq \eta, \; \forall (x,y)\in S'.$$
We encode $\eta$ by some approximation $\Tilde{\eta}$, such that $\lrabs{\Tilde{\eta}-\eta}\leq \beta$, by discretizing $[0,1]$ to $1/\beta$ buckets of size $\beta$, and $\eta$ is rounded down to the closest value $\Tilde{\eta}$. The approximation requires to encode $\log\lr{1/\beta}$ bits, and so each learner encodes $d\log\lr{1/\beta}$ bits and $d$ samples. We have $k$ weak learners, and the compression size is 
\begin{align*}
    k(d+d\log\lr{1/\beta})
    &\leq
    2kd\log\lr{1/\beta}.
\end{align*}
Therefore, we have a uniform $5\beta$-approximate compression of size
\begin{align*}
    \cO\lr{\fat\lr{\cH,\beta/8}\fat^*\lr{\cH,c\beta}\log^2\lr{\frac{\fat\lr{\cH,\beta/8}}{\beta^2}}
    \log^2\lr{\frac{\fat^*\lr{\cH,c\beta}}{\beta}}\log\frac{1}{\beta}}.
\end{align*}

\end{enumerate}
\paragraph{Generalization bound.}
Let $(\kappa,\rho)$ be the compression scheme and $\lrabs{\kappa(S)}$ the compression size. Let $\widehat{\err}_{\eta}(h;S)$ be the empirical loss of $h$ on $S$ with the $\eta$-ball robust loss.
We can upper bound the error as follows, 
\begin{align*}
{\err_{\eta}}\lr{\rho({\kappa({S})});\cD}
&\overset{(i)}{\lesssim}
\widehat{\err}_{\eta}\lr{\rho({\kappa({S})});S}
+
\frac{|\kappa(S)|\log(m)+\log \frac{1}{\delta}}{m}
\\
&\overset{(ii)}{\lesssim}
\widehat{\err}_{\eta}\lr{h^\star;S}
+5\beta
+
\frac{|\kappa(S)|\log(m)+\log \frac{1}{\delta}}{m}
\\
&\overset{(iii)}{\lesssim}
\err_{\eta}\lr{h^\star;\cD}
+5\beta
+
\frac{|\kappa(S)|\log(m)+\log \frac{1}{\delta}}{m}
+
\frac{\log \frac{1}{\delta}}{m}
\\
&\lesssim
\err_{\eta}\lr{h^\star;\cD}
+5\beta
+
\frac{|\kappa(S)|\log(m)+\log \frac{1}{\delta}}{m},
\end{align*}
(i) follows from a generalization of sample compression scheme in the realizable case, see \cref{lem:sample-compression},
(ii) follows from the approximate sample compression we proved above,
(iii) follows from Hoeffding's inequality.
We take $m$ sufficiently large such that
\begin{align*}
\frac{|\kappa(S)|\log(m)+\log \frac{1}{\delta}}{m}
\lesssim
\epsilon.
\end{align*}
By plugging it in the compression size and re-scaling $\beta$, we get a sample complexity of size
\begin{align*}
\cM
=
\cO\lr{
    \frac{1}{\epsilon}\lr{
\fat\lr{\cH,c\beta}\fat^*\lr{\cH,c\beta}\log^2\lr{\frac{\fat\lr{\cH,c\beta}}{\beta^2}}
    \log^2\lr{\frac{\fat^*\lr{\cH,c\beta}}{\beta^2}}\log\frac{1}{\beta}\log\frac{1}{\epsilon}+\log\frac{1}{\delta}
}
},
\end{align*}
for some numerical constant $c\in (0,\infty)$.
\end{proof}

\subsection{Agnostic}
\begin{proof}[of \cref{thm:ag-uni-reg}]
The construction follows a reduction to the realizable case similar to \cite{david2016supervised}, which is for the non-robust zero-one loss. Moreover, we use a margin-based analysis of $\medboost$ algorithm (see \citet[Theorem 1]{kegl2003robust}).

 Denote $\Lambda_{\RE} = \Lambda_{\RE}(\eta, \beta,1/3,1/3,\cH,\cU)$, the sample complexity of Robust $(\eta,\beta)$-regression for a class $\cH$ with respect to a perturbation function $\cU$, taking $\epsilon=\delta=1/3$ 

Using a robust $\ERM$ in order to find a maximal subset $S'\subseteq S$ with zero empirical robust loss (for the $\eta$-ball loss),
such that $\inf_{h\in\cH}\widehat{\err}_\eta\lr{h;S'}=0$. Now, $\Lambda_{\RE}$ samples suffice for weak robust learning for any distribution $\cD$ on $S'$.

Compute the $\mathrm{MedBoost}$ on $S'$, with $T \approx \log\lr{|S'|}$ boosting rounds, where each weak robust learner is trained on $\approx\Lambda_{\RE}$ samples. The returned weighted median $\hat{h}=\med\Lr{\hat{h}_1(z),\ldots,\hat{h}_T(z);\alpha_1,\ldots,\alpha_T}$
satisfies  $\widehat{\err}_{\eta+\beta}\lr{\hat{h};S'}=0$, and each hypothesis $\hat{h}_t\in \lrset{\hat{h}_1,\ldots,\hat{h}_T}$ is representable as set of size $\cO(\Lambda_{\RE})$. This defines a compression scheme of size $\Lambda_{\RE}T$.
%

By plugging it into an agnostic sample compression bound \cref{lem:sample-compression-agnostic}, we have a sample complexity of $\Tilde{\cO}\lr{\frac{\Lambda_{\RE}}{\epsilon^2}}$, which translates into $\Tilde{\cO}\lr{\frac{\fat\lr{\cH,c\beta}\fat^*\lr{\cH,c\beta}}{\epsilon^2}}$, for some numerical constant $c\in (0,\infty)$.
\end{proof}

\subsection{Naive approach with a fixed cutoff}\label{app:naive-approach-fixed-cutoff}
An agnostic learner for robust $(\eta,\beta)$-regression does not apply to the robust regression setting. The reason is that the optimal function in $\cH$ has different scales of robustness on different points. 
we show that by using a fixed cutoff for all points we can obtain an error of $\sqrt{\text{OPT}_\cH}+\epsilon$, where
\begin{align*}
    \mathrm{OPT}_\cH
    =
    \inf_{h\in\cH}\hE_{(x,y)\sim\cD} \LRbra{\sup_{z\in \cU(x)}\lrabs{h(z)- y}}.
\end{align*}
\begin{theorem}\label{thm:ag-reg}
For any $\cH\subseteq [0,1]^\cX$ with finite $\gamma$-fat-shattering for all $\gamma>0$, any $\cU:\cX\rightarrow 2^\cX$, and any $\eta, \beta, \epsilon,\delta \in(0,1)$, for some numerical constant $c\in(0,\infty)$, with probability $1-\delta$, \cref{alg:agnostic-robust} outputs a function with error at most $\sqrt{\mathrm{OPT}_\cH}+\epsilon$, by using a sample of size
\begin{align*}
    \Tilde{\Omega}\lr{\frac{\fat\lr{\cH,c\beta}\fat^*\lr{\cH,c\beta}}{\epsilon^2}+\frac{1}{\epsilon^2}\log\frac{1}{\delta}}.
\end{align*}
\end{theorem}

\begin{algorithm}[H]
    \caption{}\label{alg:agnostic-robust}
    \textbf{Input}: $\cH\subseteq [0,1]^\cX$, $S=\lrset{\lr{x_i,y_i}}^m_{i=1}$, $\Tilde{S}=\lrset{\lr{x_i,y_i}}^n_{i=1}$.\\
    \textbf{Algorithms used:} Agnostic learner for Robust $(\eta,\epsilon)$-regression (see \cref{thm:ag-uni-reg}): $\mathrm{Agnostic\text{-}}(\eta+\epsilon)\text{-}\mathrm{Regressor}$. 
    \begin{enumerate}
        \item Define a grid $\Theta = \lrset{\frac{1}{m}, \frac{2}{m}, \frac{4}{m}, \frac{8}{m},\ldots,1}$.
        \item Define $\cH_{\Theta} = \lrset{h_{\theta}=\mathrm{Agnostic}\text{-}\theta\text{-}\mathrm{Regressor}(S):\theta\in\Theta}$.
        \item Find an optimal function on the holdout set
        \begin{align*}
            \hat{h}_\theta=\argmin_{h_\theta\in{\cH_\Theta}}
            \frac{1}{\Lrabs{\Tilde{S}}}
            \sum_{(x,y)\in \Tilde{S}}
             \hI\lrbra{\sup_{z\in\cU(x)}\lrabs{h_\theta(z)- y}\geq \theta}
        \end{align*}
    \end{enumerate}
    \textbf{Output:} $\hat{h}_{\theta}$.
\end{algorithm}

\begin{proof}[of \cref{thm:ag-reg}]
Let 
\begin{align*}
    \mathrm{OPT}_\cH
    =
    \inf_{h\in\cH}\hE_{(x,y)\sim\cD} \LRbra{\sup_{z\in \cU(x)}\lrabs{h(z)- y}},
\end{align*}
which is obtained by $h^\star\in\cH$.
By Markov Inequality we have
\begin{align*}
    \hP_{(x,y)\sim\cD}\lr{\sup_{z\in \cU(x)}\Lrabs{h^\star(z)-y} >\eta }
    \leq 
    \frac{\hE_{(x,y)\sim \cD}\lrbra{\sup_{z\in \cU(x)}\Lrabs{h^\star(z)-y}}}{\eta}.
\end{align*}
Taking $\eta=\sqrt{\mathrm{OPT}_\cH}$, 
\begin{align*}
    \hP_{(x,y)\sim\cD}\lr{\sup_{z\in \cU(x)}\Lrabs{h^\star(z)-y} >\sqrt{\mathrm{OPT}_\cH} }
    \leq 
    \frac{\mathrm{OPT}_\cH}{\sqrt{\mathrm{OPT}_\cH}}
    \\
    =
    \sqrt{\mathrm{OPT}_\cH}.
\end{align*}
This means that we can apply the algorithm for agnostic robust uniform $\eta$ regression with $\eta=\sqrt{\mathrm{OPT}_\cH}$, and obtain an error of $\sqrt{\mathrm{OPT}_\cH}+\epsilon$.
The problem is that $\mathrm{OPT}_\cH$ is not known in advance. To overcome this issue, we can have a grid search on the scale of $\eta$, and then verify our choice using a holdout training set.

We define a grid,$\Theta = \lrset{\frac{1}{m}, \frac{2}{m}, \frac{4}{m}, \frac{8}{m},\ldots,1}$, such that one of its elements satisfies $\sqrt{\mathrm{OPT}_\cH}<\hat{\theta}<2\sqrt{\mathrm{OPT}_\cH}$.

For each element in the grid, we compute the agnostic regressor for the $\eta$-robust loss. That is, we define $\cH_{\Theta} = \lrset{h_{\theta}=\mathrm{Agnostic}\text{-}\theta\text{-}\mathrm{Regressor}(S):\theta\in\Theta}$.   

We choose the optimal function on a holdout labeled set $\Tilde{S}$ of size $\approx \frac{1}{\epsilon^2}\log\frac{1}{\delta}$,
\begin{align*}
            \hat{h}_\theta=\argmin_{h_\theta\in{\cH_\Theta}}
            \frac{1}{\Lrabs{\Tilde{S}}}
            \sum_{(x,y)\in \Tilde{S}}
             \hI\lrbra{\sup_{z\in\cU(x)}\lrabs{h_\theta(z)- y}\geq \theta}.
\end{align*}
    
With high probability, the algorithm outputs a function with error at most $\sqrt{\mathrm{OPT}_\cH}+\epsilon$ for the $\ell_1$ robust loss, using a sample of size 
\begin{align*}
    \Tilde{\cO}\lr{\frac{\fat\lr{\cH,c\epsilon}\fat^*\lr{\cH,c\epsilon}}{\epsilon^2}}.
\end{align*}
\end{proof}

\end{document}